%% file: neurips_2025.tex
\documentclass{article}

\PassOptionsToPackage{numbers, compress}{natbib}
     \usepackage[preprint]{neurips_2025}

\usepackage[utf8]{inputenc} % allow utf-8 input
\usepackage[T1]{fontenc}    % use 8-bit T1 fonts
\usepackage{hyperref}       % hyperlinks
\usepackage{url}            % simple URL typesetting
\usepackage{booktabs}       % professional-quality tables
\usepackage{amsfonts}       % blackboard math symbols
\usepackage{nicefrac}       % compact symbols for 1/2, etc.
\usepackage{microtype}      % microtypography
\usepackage{xcolor}         % colors
\usepackage{amsmath}	
\usepackage{amssymb}	
\usepackage{booktabs}
\usepackage{times}
\usepackage{microtype}
\usepackage{epsfig}
\usepackage{caption}
\usepackage{float}
\usepackage{placeins}
\usepackage{color, colortbl}
\usepackage{stfloats}
\usepackage{enumitem}
\usepackage{tabularx}
\usepackage{xstring}
\usepackage{multirow}
\usepackage{xspace}
\usepackage{url}
\usepackage{xcolor}
\usepackage[hang,flushmargin]{footmisc}
\usepackage{soul}
\usepackage{makecell} 
\usepackage{wrapfig,lipsum,booktabs}
\definecolor{iccvblue}{rgb}{0.21,0.49,0.85}
\title{DynaNav: Dynamic Feature and Layer Selection for Efficient Visual Navigation}

\author{
Jiahui~Wang$^{1}$ ~~~ 
Changhao~Chen$^{2}$ ~~~\\
$^1$College of Design and Engineering, National University of Singapore\\
$^2$PEAK-Lab, The Hong Kong University of Science and Technology (Guangzhou)\\
{\texttt{wjiahui@u.nus.edu}}~~~~~
{\texttt{changhaochen@hkust-gz.edu.cn}}
}

\begin{document}

\maketitle
\input{sec/0_abstract}

\input{sec/1_intro}
\input{sec/2_related}
\input{sec/3_method}
\input{sec/4_experiment}

\input{sec/5_conclusion}
\input{sec/6_acknowledgement}

{
    \small
    \bibliographystyle{IEEEtran}
    \bibliography{neurips_2025.bib}
}
\input{sec/checklist}
\clearpage \appendix \input{sec/appendix}

\end{document}

%% file: sec/0_abstract.tex
\begin{abstract}
%Recent advancements in visual navigation, such as ViNT, have demonstrated the effectiveness of transformer decoders in this domain. However, these methods face two critical challenges: (1) their high computational cost hinders real-time deployment, and (2) their lack of interpretability reduces reliability in practical applications. In this work, we are the first to introduce a dynamic early exit strategy for visual navigation models, which adaptively activates transformer decoder layers based on input complexity, significantly improving computational efficiency without compromising performance. Additionally, we introduce a novel dynamic hard feature selector to filter out redundant information of input features. By sparslizing the features, it not only enhances the robustness of encoded features and improves interpretability but also increases the success rate and stability of early exits. Through a Bayesian Optimization process after pre-training, our approach reduces computational cost by $2.26\times$ FLOPs compared to conventional methods. Importantly, rather than developing an entirely new navigation framework, our objective is to enhance the efficiency and reliability of existing architectures. Comprehensive evaluations on standard benchmarks and CARLA simulations demonstrate that our method not only maintains competitive navigation performance but also achieves superior computational efficiency, making real-time deployment more feasible for real-world applications.
Visual navigation is essential for robotics and embodied AI. However, existing foundation models, particularly those with transformer decoders, suffer from high computational overhead and lack interpretability, limiting their deployment in resource-tight scenarios.
To address this, we propose \textbf{DynaNav}, a \textbf{Dyna}mic Visual \textbf{Nav}igation framework that adapts feature and layer selection based on scene complexity. It employs a trainable hard feature selector for sparse operations, enhancing efficiency and interpretability. Additionally, we integrate feature selection into an early-exit mechanism, with Bayesian Optimization determining optimal exit thresholds to reduce computational cost.
Extensive experiments in real-world-based datasets and simulated environments demonstrate the effectiveness of DynaNav. Compared to ViNT, DynaNav achieves a $2.26\times$ reduction in FLOPs, 42.3\% lower inference time, and 32.8\% lower memory usage, while improving navigation performance across four public datasets. 
\end{abstract}

%% file: sec/1_intro.tex
% \vspace{-3mm}
\section{Introduction}
\label{sec:intro}
% \vspace{-2mm}
Visual navigation is a fundamental capability for robotics and embodied AI, enabling autonomous agents to perceive, interpret, and navigate complex 3D environments based on visual inputs~\cite{zhang2024vision,li2025human,bar2024navigationworldmodels}. Its applications span real-world scenarios, such as delivery and logistics, as well as virtual domains, including gaming and simulation. By bridging perception and action, visual navigation plays a crucial role in intelligent systems.
Recently, there has been growing interest in developing foundation models for visual navigation~\cite{shah2023gnm,shah2023vint,sridhar2023nomad,shah2021ving,shah2023lm,cai2024bridging,zhou2024navgpt,zhang2024vision}. ViNT~\cite{shah2023vint} is a notable example that learns from large-scale egocentric observations using transformer layers on CNN-extracted features, demonstrating strong generalization across robotic platforms and environments. NoMad~\cite{sridhar2023nomad} further builds on this by incorporating a diffusion policy and a goal-masking mechanism. PixNav~\cite{cai2024bridging} utilizes textual heuristics and large language models(LLMs) to explore zero-shot possibility. However, these approaches, particularly those relying on deep neural architectures such as transformer decoders, introduce significant computational overhead, posing challenges for edge deployment where efficiency is paramount.

Robotic applications demand greater efficiency than large cloud-based models. As the trend toward efficient foundation models continues~\cite{DeeR-VLA, sun2023dime}, reducing the computational burden of visual navigation models is a key challenge. Additionally, existing models function as "black boxes," raising concerns about interpretability. As humans and intelligent agents increasingly coexist, explainability becomes essential. These challenges lead to two critical research questions:
% \vspace{-2mm}
\begin{itemize}
    \item \emph{Is it necessary to activate all transformer layers for every navigation scenario?}
    \item \emph{Which features are most important in the decoding process, and can we identify the most salient regions or pixels for navigation?}
\end{itemize}
% \vspace{-2mm}
Humans do not always activate all neurons for visual tasks~\cite{bharath2008next}; rather, the brain dynamically recruits resources based on task complexity. Inspired by this, we propose that visual navigation models should adopt dynamic inference mechanisms, selectively utilizing features and neural layers based on scene complexity. In simpler scenarios, the model should rely on fewer features and layers for efficient computation, whereas in more complex tasks, it should allocate additional resources to ensure accurate decision-making.

To this end, we propose \textbf{DynaNav}, a highly efficient \textbf{Dyna}mic Visual \textbf{Nav}igation framework that adaptively selects relevant features and neural layers based on visual observations. Our approach employs a trainable hard feature selector to create sparse representations, enabling computationally efficient sparse operations at the feature level.
% \textcolor{red}{This dynamic feature-masking not only enhances interpretability but also improves feature robustness.}
{This dynamic feature masking not only lowers computational overhead but also improves the understanding of which regions more relevantly influence the inference of visual navigation models, thereby enhancing explainability.} Additionally, we introduce an early-exit strategy for deep Transformer layers by integrating feature selection into the early-exit mechanism, improving stability and computational efficiency. After training the decoder, Bayesian Optimization determines optimal early-exit thresholds. During inference, if a layer's feature meets its threshold, computation terminates early, significantly reducing overall computational cost.
{Extensive experiments on real-world datasets and in simulated environments demonstrate the effectiveness of our proposed DynaNav.} Compared to ViNT~\cite{shah2023vint}, DynaNav achieves a $2.26\times$ reduction in FLOPs, 42.3\% lower inference time, and 32.8\% lower memory usage while improving navigation performance across four public datasets. 
To the best of our knowledge, this is the first work to introduce dynamic network mechanisms to visual navigation models. To sum up, the main contributions of our work can be summarized as follows:
% \vspace{-1mm}
\begin{itemize}
    \item We propose DynaNav, a highly efficient and effective dynamic neural model for visual navigation, introducing a novel feature and layer selection strategy to improve efficiency without compromising performance.

    \item We integrate sparse feature selection into the early exit mechanism, improving the stability and success rate of dynamic layer inference, while the visualized mask enhances the interpretability of the navigation decision process.

    \item Extensive experiments and simulations demonstrate that DynaNav achieves more than twice the efficiency while maintaining comparable success rates.
\end{itemize}

%% file: sec/2_related.tex
% \vspace{-3mm}
\section{Related Work}
% \vspace{-3mm}
\label{sec:related}
\subsection{End-to-end Visual Navigation}
% \vspace{-2mm}
Nowadays, conducting robot learning from diverse datasets to obtain a general model is becoming more and more popular~\cite{devin2017learning,dasari2020robonet,yu2020bdd100k}. Nonetheless, current approaches rely on real-world data, which is usually {costly to obtain}, lacks generalization, and is highly coupled with specific robot settings that are hard to transfer to different platforms~\cite{kadian2020sim2real,anderson2021sim}. Instead, our paper follows the paradigm of learning navigation behavior from data collected across multiple different real-world robotic systems~\cite{hirose2023exaug,loquercio2018dronet,shah2023gnm} while focusing on training a foundation model that can be adapted for various downstream tasks in zero-shot or with small amounts of data. To this end, models like RT-1, I2O, and GNM~\cite{truong2024indoorsim,shah2023gnm,brohan2022rt} provide useful insights that study broad generalization across environments and embodiments for robots deployed in real-world settings. GNM~\cite{shah2023gnm} demonstrates policy learning from heterogeneous RGB datasets but focuses on the singular task of reaching image goals in the zero-shot setting. ViNT~\cite{shah2023vint} trains an effective visual navigation policy that can solve a range of downstream tasks, such as navigating to GPS goals~\cite{savva2019habitat}, goal images~\cite{zhu2017target}, and skill-conditioned driving~\cite{codevilla2018end}. Building upon extensive prior work in visual navigation, ViNT combines two key elements: it uses topological graphs to keep track of how spaces are connected in the environment while employing trained policies to handle the detailed movement controls~\cite{savinov2018semi,bruce2018learning,faust2018prm,meng2020scaling,shah2021ving,hirose2019deep}. Recently, NoMaD~\cite{sridhar2023nomad} boosted the navigation task in previously unseen environments with goal masking techniques and diffusion policy. 
\subsection{Dynamic Network and Early Exit}
% \vspace{-2mm}
Dynamic networks~\cite{han2021dynamic,del2023skipdecode,raposo2024mixture} tend to optimize models that can modify their architecture or parameters based on the input during the inference process. There are many techniques to achieve a dynamic network depth-wise and width-wise. For instance, layer-skipping~\cite{elhoushi2024layerskip}, neurons-skipping~\cite{bengio2013estimating}, and low rank approximation (LoRA)~\cite{davis2013lora}. Moreover, some dynamic networks focus on adjusting the shape and value of weights adaptively during the inference, such as deformable convolution~\cite{dai2017deformable}, dynamic filter network~\cite{jia2016dynamic}, etc.. Among these techniques, early exiting gained popularity because of the prevalent Transformer-based model, which fits the inherent architecture with stacked blocks.

\noindent\textbf{Early exiting} is a depth-wise dynamic method for halting forward propagation at a certain layer based on intermediate predictions. This technique has been well explored in both computer vision~\cite{wang2021not,bolukbasi2017adaptive,msdnet,han2023dynamic,yang2021condensenet,10508473}, language processing~\cite{elbayad2019depth,xin2021berxit,xin2020deebert,liu2020fastbert,mangrulkar2022be3r,chen2023ee,schuster2022confident}, and multimodality~\cite{fei2022deecap,tang2023you}. One challenge in implementing early-exiting models lies in finding a suitable metric to decide when to make an intermediate prediction. Commonly, metrics like probability confidence~\cite{msdnet} or entropy value~\cite{xin2020deebert} are employed in traditional vision tasks. Some research also pointed out the possibility of using learning-based early exit, which relies on a trained network~\cite{ghodrati2021frameexit,xin2021berxit,Ni2024AdaNAT,fang2024real}. Recent research~\cite{del2023skipdecode,elhoushi2024layerskip} has extended early exiting to the LLMs, which treat the autoregressive task as a classification subgoal.

In this work, we are the first to utilize the idea of early exiting on an end-to-end visual navigation model. We further develop the current method~\cite{DeeR-VLA} with the integration of sparse feature selection to the Bayesian optimization process in obtaining the desirable metric.
\input{figs/pipeline}

%% file: figs/pipeline.tex
\begin{figure*}
    \centering
    \includegraphics[width=1\linewidth]{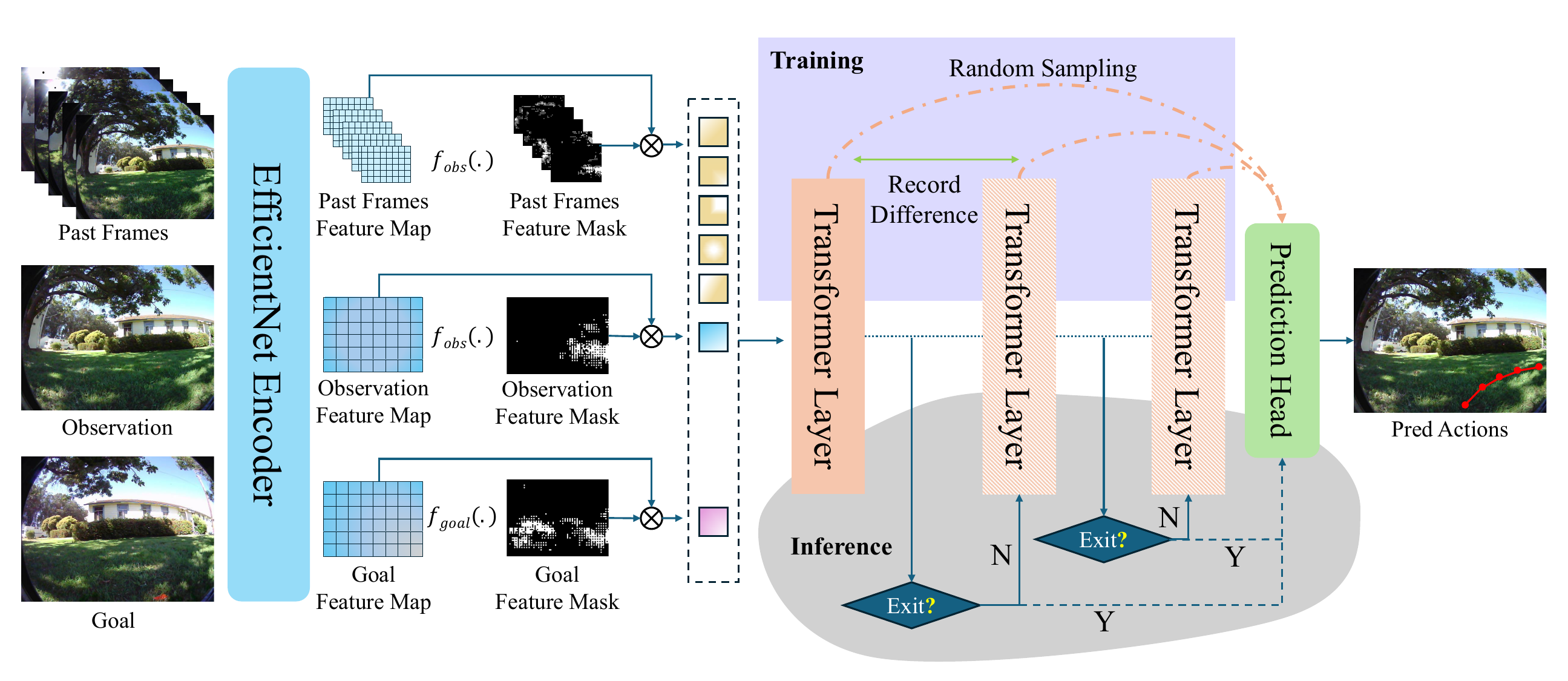}
    \caption{The architecture of our DynaNav framework. DynaNav employs two encoder instances of the same architecture: one processes the concatenated observation and historical frame sequence, while the other extracts features from the early-fused combination of the current observation and goal image. Two independent feature selectors generate masks for observations and the goal, which are then tokenized and fed into a Transformer decoder. Training incorporates stochastic early-exit triggers at intermediate decoder layers. During inference, a decision step at each layer evaluates optimized thresholds to determine whether early exit conditions are met.}
    \label{fig:pipeline}
    % \vspace{-5mm}
\end{figure*}

%% file: sec/3_method.tex
\section{Dynamic Visual Navigation Model}
% \vspace{-2mm}
\label{sec:method}
\subsection{Framework Overview}
% \vspace{-2mm}
To enable efficient and effective visual navigation, we propose DynaNav, a dynamic navigation pipeline illustrated in Figure~\ref{fig:pipeline}. Unlike previous end-to-end models with static network inference, DynaNav integrates a dynamic feature selector and an early-exit mechanism, reducing computational costs while enhancing explainability and robustness.
DynaNav begins with an EfficientNet backbone~\cite{tan2019efficientnet} to extract features from RGB image sequences. Building on ViNT~\cite{shah2023vint}, we introduce a feature selector module that generates masks before feature processing in the Transformer decoder, allowing for sparse computation. Additionally, we implement a dynamic Transformer decoder, enabling predictions at intermediate layers to improve efficiency. Finally, Bayesian optimization~\cite{DeeR-VLA} determines the optimal early-exit thresholds for our jointly trained model, further minimizing computational overhead.
% These masks filter out irrelevant information, leading to more robust representations. We observe that ViNT’s~\cite{shah2023vint} indiscriminate decoding introduces computational redundancy: While complex tasks (where the observation and goal differ significantly) may require full decoder processing, simpler tasks (where they are similar) can achieve accurate predictions with fewer layers. Based on this insight, 

\textbf{Feature Extraction.} 
% We selected EfficientNet-B0 as our visual encoder for its innovative compound scaling method, which optimally balances network width, depth, and resolution simultaneously. Unlike traditional arbitrary scaling approaches that can lead to diminishing returns, this method maintains efficiency as model capacity increases. Furthermore, EfficientNet-B0 demonstrates superior performance metrics compared to models of similar size. Mathematically, it takes the input consisting of consecutive visual observations  $o_i, \ i\in[t-p,t]$ and a goal image $o_s$. For each visual observation, it is first projected to the latent space by the encoder and denoted as $\psi(o_i) \in \mathbb{R}^{H\times W \times C}$, where $\psi(\cdot)$ is the network, $H,W, C$ are corresponding height, width and channel number of the feature map. Especially for the goal image, the early fusion strategy is adopted to strengthen the connection between the current observation and the goal. Its feature is also extracted by an individual efficientnet and denoted as $\phi([o_t;o_s]) \in \mathbb{R}^{H\times W \times C}$, where $[;]$ is the concatenation operation. The detailed hyperparameter setting is illustrated in Section~\ref{sec:app_hyperparam}.
We chose EfficientNet-B0 as our visual encoder due to its innovative compound scaling method, which optimally balances network width, depth, and resolution. Mathematically, the encoder processes an input sequence consisting of consecutive visual observations \( \mathbf{o}_i \), where \( i \in [t-p, t] \), along with a goal image \( \mathbf{o}_s \). Each observation is first mapped to a latent space representation by the encoder, denoted as \( \psi(\mathbf{o}_i) \in \mathbb{R}^{H \times W \times C} \), where \( \psi(\cdot) \) represents the network, and \( H, W, C \) correspond to the height, width, and number of channels in the feature map. To enhance the connection between the current observation and the goal, we adopt an early fusion strategy. Specifically, the goal image is processed separately by another EfficientNet instance, producing a feature representation denoted as \( \phi([\mathbf{o}_t; \mathbf{o}_s]) \in \mathbb{R}^{H \times W \times C} \), where \([;]\) represents concatenation. The detailed hyperparameter settings are provided in Section~\ref{sec:app_hyperparam}.

\textbf{Dynamic Feature Selector.} However, when the embedded feature map is large~\cite{DeeR-VLA}, computing such a tensor in a transformer incurs significant computational costs~\cite{attention}, which limits the navigation model's efficiency. Moreover, not all pixels in the observations and goal are essential; some redundant pixels can be ignored to improve processing efficiency~\cite{he2022masked}. To address this, we introduce a dynamic hard feature selector that generates a mask to filter out pixels with minimal relevance to the final prediction. 
% The mask is then applied directly to the observations, setting irrelevant features to zero. This not only enables sparse computation and storage but also enhances the model’s robustness by encouraging it to learn more generalizable features rather than focusing on specific regions or objects in the image.  

\textbf{Transformer Decoder.} 
% A transformer decoder $D$ is then utilized to obtain the contextual feature for action prediction, which is depicted in Figure~\ref{fig:pipeline}. The continuous stacked multi-head self-attention (MHSA) layers are utilized to learn the contextual information of the visual token. Mathematically, we define the intermediate token as:
% \begin{equation}
%     x_i = D_{1:i}\left([m_{t-p:t}*\psi(o_{t-p:t});m_s*\phi([o_t;o_s])]\right),
% \end{equation}
% where $m_{t-p:t}$ represent the generated masks for $o_i,i\in[t-p,t]$, $D_{1:i}, i \in [1,l]$ represent the $i$ continuos layers. The detailed process of obtaining $m$ is denoted in Section~\ref{sec:fs}.
After the feature selection process, a transformer decoder \( \mathbf{D} \) is employed to extract contextual features for action prediction, as illustrated in Figure~\ref{fig:pipeline}. The stacked multi-head self-attention (MHSA) layers continuously refine the contextual information of visual tokens. Formally, we define the intermediate token as:
% \vspace{-1mm}
\begingroup
\begin{equation}
    \mathbf{x}_i = D_{1:i}\left([\mathbf{m}_{t-p:t}*\psi(\mathbf{o}_{t-p:t});\mathbf{m}_s*\phi([\mathbf{o}_t;\mathbf{o}_s])]\right),
\end{equation}
\endgroup
where \( \mathbf{m}_{t-p:t} \) represents the generated masks for \(\mathbf{o}_i \), \( i \in [t-p,t] \), and \( D_{1:i} \), \( i \in [1,l] \), denotes the first \( i \) layers of the decoder. The process for obtaining \( \mathbf{m} \) is detailed in Section~\ref{sec:fs}. 

\textbf{Navigational Action Prediction.}   Finally, a head network is trained to predict both the action \(\mathbf{a}_t\) and the waypoint distance vector \(\mathbf{w}_t\). When feature selection is applied, this prediction process can be formulated as \(\mathbf{a}_t, \mathbf{w}_t = h(\mathbf{x}_l)\), where \(h\) represents the prediction head. In our implementation, \(h\) consists of a 4-layer transformer followed by an MLP with a single hidden layer.  

\textbf{Training Objective.} During training, we sample a sequence of visual images from the dataset to construct the observation \( \mathbf{o}_{t-p:t} \). A goal image \( \mathbf{o}_s \) is randomly selected for a valid prediction length, where \( \mathbf{o}_s = \mathbf{o}_{t+d} \) and \( d \in [t_{\min}, t_{\max}] \). The corresponding action sequence \( \mathbf{a}^{\text{gt}}_t = \mathbf{a}_{t:t+d} \) and waypoint \( \mathbf{w}^{\text{gt}}_t \) serve as ground truth. The objective of training is to maximize the likelihood of the predicted outputs aligning with the ground truth, formulated as:  
% \vspace{-1mm}
\begingroup
\begin{equation}
    \mathcal{L}=\mathbb{E}\left[\log p \left(\mathbf{a}^{\text{gt}}_t \mid \mathbf{a}_t\right)+\lambda \log p\left(\mathbf{w}^{\text{gt}}_t \mid \mathbf{w}_t\right)\right].
\end{equation}
\endgroup
% \vspace{-7mm}
\subsection{Dynamic Sparse Feature Selection}\label{sec:fs}
% \vspace{-2mm}
\input{figs/selector} % Current end-to-end visual navigation models are majorly black boxes~\cite{gnm,shah2023vint,shah2022viking,shah2021ving}. One problem is that for a navigation scene, we don't know which part of the observation is useful for action prediction. If we can acquire such a piece of information, many preprocessing methods can be implemented to improve the performance of the model. Moreover, we raise the question that \emph{is all pixels should be utilized indiscriminately}? Intuitively, the answer is no, since only those that are significant should be prioritized to enhance the model's robustness. In real-world scenarios, the system often encounters complex conditions. For example, when obstacles obstruct the camera's view of our navigation robot, reliance on certain pixels can lead to failures. To address these challenges, we propose a novel feature selection approach based on the Gumbel-Softmax mechanism. This method aims to dynamically prioritize critical features, improving the system's performance and adaptability under varying conditions. Besides, it also offers explainability to the navigation model.
End-to-end visual navigation models often operate as black boxes~\cite{shah2023gnm,shah2023vint,shah2022viking,shah2021ving}, making it unclear which parts of an observation contribute most to action prediction. Understanding these key elements could enable targeted preprocessing to enhance model performance. This leads to a fundamental question: \emph{should all pixels be treated equally?} Intuitively, the answer is no—emphasizing only relevant pixels improves robustness. In real-world scenarios, indiscriminate reliance on all pixels can lead to failures, especially when obstacles obstruct a robot’s camera. To address this, we introduce a feature selection approach based on the Gumbel-Softmax mechanism~\cite{gumblesoftmax}, dynamically prioritizing critical features. {This improves performance and adaptability across diverse environments and provides meaningful insights into the model's region of interest.} The feature selector functions as a classification network, assigning each pixel a probability score to generate masks for different input features. As shown in Figure~\ref{fig:selector}, the feature selector \( f(\cdot) \) takes encoded features as input and outputs corresponding masks as follows,
% \vspace{-1mm}
\begingroup
\begin{equation}
        \mathbf{m}_i = f(\psi(\mathbf{o}_i)); \ \ 
        \mathbf{m}_s = f(\phi([\mathbf{o}_t;\mathbf{o}_s])) \in \mathbb{R}^{H\times W}.
\label{eq:fs}
\end{equation}  
\endgroup

Within the feature selector, the latent feature \( \psi(\mathbf{o}_i) \) is projected into a higher-dimensional space:
\begingroup
\begin{equation}
    \mathbf{Z}_i = MLP(\psi(\mathbf{o}_i)) \in \mathbb{R}^{H\times W\times C \times 2},
\end{equation}  
\endgroup
where \( MLP(\cdot) \) denotes a multi-layer perceptron. Here, \( z^{n, c, k}_i \) represents the unnormalized log probability of the \( k \)-th category for the \( n \)-th pixel and \( c \)-th channel. To obtain the one-hot mask, we utilize the Gumble-SoftMax trick, which first adds a log term to each element and then conducts the SoftMax. The logarithm term is defined as:
\begin{equation}
    g^{n, c, k}_i=-\log \left(-\log \left(u^{n, c, k}\right)\right); \quad u^{n, c, k} \sim {U}(0,1),
\end{equation}
and the processed value in $\mathbf{Z}_i$ is $\bar{z}^{n, c, k}_i=z^{n, c, k}_i+g^{n, c, k}_i$. Then the SoftMax is applied on $\mathbf{Z}_i$ along the last dimension:
% \vspace{-1mm}
% \begingroup
% \setlength{\belowdisplayskip}{1mm}
\begin{equation}
\hat{z}^{n, c, k}_i=\frac{\exp (\frac{\bar{z}^{n, c, k}_i}{\tau})}{\sum_{k^{\prime}=1}^2 \exp (\frac{\bar{z}^{n, c, k^{\prime}}_i}{\tau})}, \quad k=1,2,
\end{equation}
% \endgroup
where $\tau$ is a temperature hyperparameter. At last, we manually define the last dimension of $\hat{\mathbf{Z}}_i$ as the generated mask, i.e. $m_i^{n,c} = \hat{z}_i^{n,c,2}$. The feature selector will gradually filter out the undesired features as the training continues. 

\input{figs/saliency} Figure~\ref{fig:saliency} presents the visualized input and its gradients that are processed through the feature selector. The spatial importance weights are visualized through saliency maps, where the attention mask is upsampled to match the input dimensions. The brightness intensity of each pixel in the visualization corresponds to its selection probability by the feature selection mechanism. The results indicate redundancy within the input data, while the navigation model does not specifically focus on the largest common object between the observation and the goal. This finding not only supports the feasibility of filtering pixels but also enhances the interpretability of the navigation process. After selection, we can utilize data sparselization techniques~\cite{wheatman2018packed,ruiter2024value} to save space.
% Hard masks offer several advantages over soft masks in deep learning. First, hard masks contain only binary values (0 and 1) that can be stored as individual bits, while soft masks require 32-bit floating-point values. This results in hard masks requiring approximately 1/32 of the storage space compared to soft masks at the same scale. Second, the binary nature of hard masks allows matrix multiplication to be simplified into a selection operation—calculations are skipped when the mask is 0, and values are used directly when the mask is 1. In contrast, soft masks require floating-point multiplication for each element. Finally, hard masks enable the construction of efficient sparse matrix formats like CSR~\cite{wheatman2018packed} or CSC~\cite{ruiter2024value}, further improving computational efficiency. To incorporate with the early-exiting transformer decoder, we additionally utilize the number of valid pixels as an early exit threshold, which is discussed in Section~\ref{sec:bo}.

% \vspace{-3mm}
\subsection{Dynamic Transformer Layer Inference}
% \vspace{-2mm}
\subsubsection{Feature-Aware Early Exit Strategy}
Transformer-based decoders are highly effective in visual navigation, leveraging long-range dependencies and flexible adaptation~\cite{vit,awadalla2023openflamingo,liu2024llava,oquab2024dinov2}. However, models like ViNT~\cite{shah2023vint} employ a scene-agnostic decoder that activates all layers indiscriminately, disregarding scene complexity and task requirements. While beneficial for large-scale training, this approach imposes excessive computational demands on edge devices. We argue that activating every layer is often unnecessary—similar to how humans selectively engage neurons for cognitive tasks~\cite{bharath2008next,ramamurthy2015human}.
To address this, we propose a \textbf{\textit{dynamic}} navigation decoder with an early-exit mechanism, allowing the model to halt computation based on scene complexity and navigation needs. By leveraging intermediate features for action prediction, this method—the first to introduce early exiting in visual navigation—eliminates redundant computations. Additionally, we enhance efficiency and robustness by integrating feature selection as an initial step in the early-exit strategy. Our approach significantly reduces computational overhead while maintaining performance, making it well-suited for resource-constrained deployment.

% To achieve such an early-exiting strategy, Figure~\ref{fig:pipeline} illustrates the workflow of our method. Generally speaking, an early-exiting metric is applied for each decoder layer, and if the requirement is satisfied, the forward passing will be terminated and output the feature of the corresponding layer. {The previous early-exiting usually constrains the difference between features. In other words, if the variety between two features is smaller than the threshold, the early-exiting will be triggered, i.e.,
% \begin{equation}
%       \| \mathbf{x}_{i} - \mathbf{x}_{i-1} \|_2 \leq \eta_{i}, \quad \forall i \in \{1, 2, \dots, l\},
% \end{equation}
% where $\eta_i$ is the threshold and $l$ is the total length of the feature blocks.}

% DeeR-VLA~\cite{DeeR-VLA} introduces a new metric on robotic-related tasks. They argue that the feature from different transformer layers is inherently distinct, therefore, they proposed an action consistency condition that measures the action output from different layers with an action head,
% \begin{equation}
%       \| h(\mathbf{x}_{i}) - h(\mathbf{x}_{i-1}) \|_2 \leq \eta_{i}, \quad \forall i \in \{1, 2, \dots, l\}.
% \end{equation}
% However, it continues to require the activation of multiple transformer layers, which does not fully minimize the computation burden. 
Figure~\ref{fig:pipeline} outlines our early-exiting strategy workflow. 
% Typically, an early-exit metric is applied at each decoder layer, terminating the forward pass and outputting the layer’s feature if a predefined criterion is satisfied. Traditional methods, such as those in~\cite{han2021dynamic, han2023dynamic, rahmath2024early}, trigger early exiting when the L2 norm of the difference between consecutive layer features drops below a threshold, $\eta_i$, as defined by:
% \vspace{-1mm}
% \begingroup
% \setlength{\belowdisplayskip}{1mm}
% \begin{equation}
% | \mathbf{x}_{i} - \mathbf{x}_{i-1} |_2 \leq \eta_{i}, \quad \forall i \in \{1, 2, \dots, l\},
% \end{equation}
% \endgroup
% where $l$ is the total number of layers. In contrast, 
DeeR-VLA~\cite{DeeR-VLA} proposes a metric for robotic tasks, arguing that transformer layer features are inherently distinct. It uses an action consistency condition, measuring the difference in action outputs from an action head $h$:
% \vspace{-1mm}
\begingroup
\begin{equation}
| h(\mathbf{x}_{i}) - h(\mathbf{x}_{i-1}) |_2 \leq \eta_{i}, \quad \forall i \in \{1, 2, \dots, l\}.
\end{equation}
\endgroup
However, this still requires activating multiple layers, limiting computational savings. To improve efficiency, we introduce an aggressive early-exit strategy. {When the L2 difference between the goal state and the current observation falls below a training-derived threshold (based on masked pixel counts), we bypass the transformer decoder entirely and compute actions directly from the encoded tokens and a prediction head.}

% \vspace{-3mm}
\subsubsection{Adaptive Threshold Optimization}
\label{sec:bo}
% \vspace{-2mm}
% To obtain the optimal early-exiting threshold, we utilize Bayesian Optimization~\cite{shahriari2015bayes} to search for the desired value based on given constraints iteratively. Considering the final predicted action \( \mathbf{a}_t \) and waypoint \( \mathbf{w}_t \) with the ground truth \( \mathbf{a}_t^{\text{gt}} \) and \( \mathbf{w}_t^{\text{gt}} \). Our goal is to maximize the cosine similarity between prediction and ground truth by optimizing the threshold of early-exiting \( \eta = \{\eta_1, \eta_2, \dots, \eta_N\} \). Therefore, the objective function can be written as:
% \[
% \max_{\eta} J(\eta) = \frac{1}{T} \sum_{t=1}^T \Big( \text{Sim}(\mathbf{a}_t, \mathbf{a}_t^{\text{gt}}) + \lambda \cdot \text{Sim}(\mathbf{w}_t, \mathbf{w}_t^{\text{gt}}) \Big),
% \]
% where \( \text{Sim}(\mathbf{u}, \mathbf{v}) = \frac{\mathbf{u} \cdot \mathbf{v}}{\|\mathbf{u}\| \|\mathbf{v}\|} \) is the cosine similarity of two vectors,
% \( \lambda > 0 \) a hyper-weight to balance the waypoint prediction and action prediction, and 
% \( T \) is the total step number of the task.
To determine the optimal early-exit threshold, we employ Bayesian Optimization~\cite{shahriari2015bayes,DeeR-VLA} to iteratively search for the best value under given constraints. Specifically, we consider the predicted action \( \mathbf{a}_t \) and waypoint \( \mathbf{w}_t \) alongside their respective ground truth values, \( \mathbf{a}_t^{\text{gt}} \) and \( \mathbf{w}_t^{\text{gt}} \). Our objective is to maximize the cosine similarity between predictions and ground truth by optimizing the early-exiting thresholds, denoted as \( \eta = \{\eta_1, \eta_2, \dots, \eta_N\} \). Consequently, the objective function is formulated as: 
% \vspace{-2mm}
\begingroup
\begin{equation}
    \max_{\eta} J(\eta) = \frac{1}{T} \sum_{t=1}^T \Big( \text{Sim}(\mathbf{a}_t, \mathbf{a}_t^{\text{gt}};\eta) + \lambda \cdot \text{Sim}(\mathbf{w}_t, \mathbf{w}_t^{\text{gt}};\eta) \Big),
\end{equation}
\endgroup
{where \( \text{Sim}(\mathbf{u}, \mathbf{v};\eta)\) represents the cosine similarity between two vectors with a given early exit threshold $\eta$.} \( \lambda > 0 \) is a hyperparameter that balances waypoint and action prediction, and \( T \) is the total number of time steps in the task. To optimize this objective function, we introduce a penalty function \( P(\eta) \) that enforces the required constraints. This function assigns a positive value when \( \eta \) violates any constraint and remains zero otherwise. Incorporating this penalty into the optimization framework, we reformulate the problem as:  
% \vspace{-1mm}
\begingroup
\begin{equation}
    \max_{\eta} V(\eta) = J(\eta) - P(\eta),
\end{equation}  
\endgroup
where the penalty term, $ P(\eta) = \sum_{k} \xi_k \cdot \max(0, g_k(\eta))$, captures the weighted sum of constraint violations. Here, \( g_k(\eta) \) quantifies the extent to which the \( k \)-th constraint is violated, while \( \xi_k \) represents its associated weight {(remain constant)}. The specific constraints that the model must satisfy are outlined below.

% \textbf{Inference time constraint.} Assume \( \text{Time}(\eta) \) is the average inference time of the test set; we aim to make sure that it is smaller than \( T_{\max} \):
%    \begin{equation}
%           \text{Time}(\eta) = \frac{1}{T} \sum_{t=1}^T \text{Time}_t(\eta) \  \text{s.t.} \space\space \text{Time}(\eta) \leq T_{\max}
%    \end{equation}
\textbf{Inference Time Constraint}. Let \( \text{Time}(\eta) \) denote the average inference time over the entire test set, where \( \eta \) represents the early exit decision parameters. To ensure the efficiency of the network, we impose a constraint that the average inference time remains below a predefined threshold \( T_{\max} \). Mathematically, this can be formulated as:
% \vspace{-2mm}
\begingroup
\begin{equation}
    \text{Time}(\eta) = \frac{1}{T} \sum_{t=1}^{T} \text{Time}_t(\eta), \  \text{s.t.} \  \text{Time}(\eta) \leq \mathcal{T}_{\max}.
\end{equation}  
\endgroup
\( T \) is the total number of test samples, {$\mathcal{T}_{\max}$ is the maximum time}, and \( \text{Time}_t(\eta) \) denotes the inference time for the \( t \)-th sample under the given early exit strategy. This constraint ensures that the optimization selects an early exit criterion that balances computational efficiency, predictive performance, and real-time or application-specific latency requirements.

% \textbf{GPU memory constraint.}   We define \( \text{Mem}(\eta) \) as the average usage of GPU memory, and it is supposed to be not greater than \( G_{\max} \):
%    \begin{equation}
%        \text{Mem}(\eta) = \max_{t=1,\dots,T} \text{Mem}_t(\eta) \space\space \text{s.t.} \  \text{Mem}(\eta) \leq G_{\max}
%    \end{equation}
\textbf{GPU Memory Constraint}.To ensure efficient deployment under limited GPU resources, we define \( \text{Mem}(\eta) \) as the GPU memory usage when applying the early exit strategy. Since memory consumption can fluctuate during inference, we consider the peak memory usage across all inference steps and enforce an upper bound constraint:  
\begingroup
\begin{equation}
    \text{Mem}(\eta) = \max_{t=1,\dots,T} \text{Mem}_t(\eta), \  \text{s.t.} \  \text{Mem}(\eta) \leq G_{\max}
\end{equation}  
\endgroup
where \( \text{Mem}_t(\eta) \) represents the memory consumption at time step \( t \), and \( G_{\max} \) denotes the maximum allowable GPU memory. This constraint ensures that Bayesian optimization selects an early exit criterion that not only improves efficiency but also maintains feasibility within hardware limitations.
   
% \textbf{FLOPs constraint.} For the FLOPs in the transformer decoder, we use  \( \text{FLOPs}(\eta) \) as the average FLOPs, which is designed to be less than \( F_{\max} \):
%     \begin{equation}
%         \text{FLOPs}(\eta) = \max_{t=1,\dots,T} \text{FLOPs}_t(\eta) \space\space \text{s.t.} \  \text{FLOPs}(\eta) \leq F_{\max}
%     \end{equation}
\textbf{FLOPs Constraint}. One of the most critical considerations in optimizing the early exit strategy is controlling the computational cost, particularly in the transformer decoder. To achieve this, we define \( \text{FLOPs}(\eta) \) as the average floating point operations (FLOPs) required per trajectory. Our goal is to ensure that the computational complexity remains within a predefined upper bound, \( F_{\max} \), while maintaining the model’s performance. Formally, we express this constraint as:
\begingroup
\begin{equation}
    \text{FLOPs}(\eta) = \frac{1}{T} \sum_{t=1}^{T} \text{FLOPs}_t(\eta) \  \text{s.t.} \  \text{FLOPs}(\eta) \leq F_{\max}.
\end{equation}
\endgroup
Here, \( \text{FLOPs}_t(\eta) \) represents the computational cost at each exit point \( t \).  By integrating this constraint into our Bayesian optimization framework, we explore the trade-off between computational efficiency and model accuracy, enabling us to identify the most effective early exit criteria within the computational limits.

% \textbf{State Initialization.} To determine the optimal threshold for early exiting, we employ online learning algorithms that iteratively refine thresholds based on similarity function feedback. While DeeR-VLA~\cite{DeeR-VLA} utilizes empirical initialization followed by Bayesian optimization, this simple initialization approach reduces optimization efficiency by requiring additional iterations for convergence. To address this limitation, we enhance the initialization process by incorporating the count of unmasked features as a prior. The intuition behind this approach is that a low number of unmasked features suggests the model perceives the navigation scene as relatively simple. Conversely, when more features are selected and processed by the decoder, this indicates the navigation task has higher complexity. Therefore, we allocate a broader search space for challenging scenarios while employing more stringent metrics for relatively simple navigation scenes. 

%% file: figs/selector.tex
\begin{wrapfigure}{r}{0.55\linewidth}
    \centering
    \vspace{-11mm}
    \includegraphics[width=1.0\linewidth]{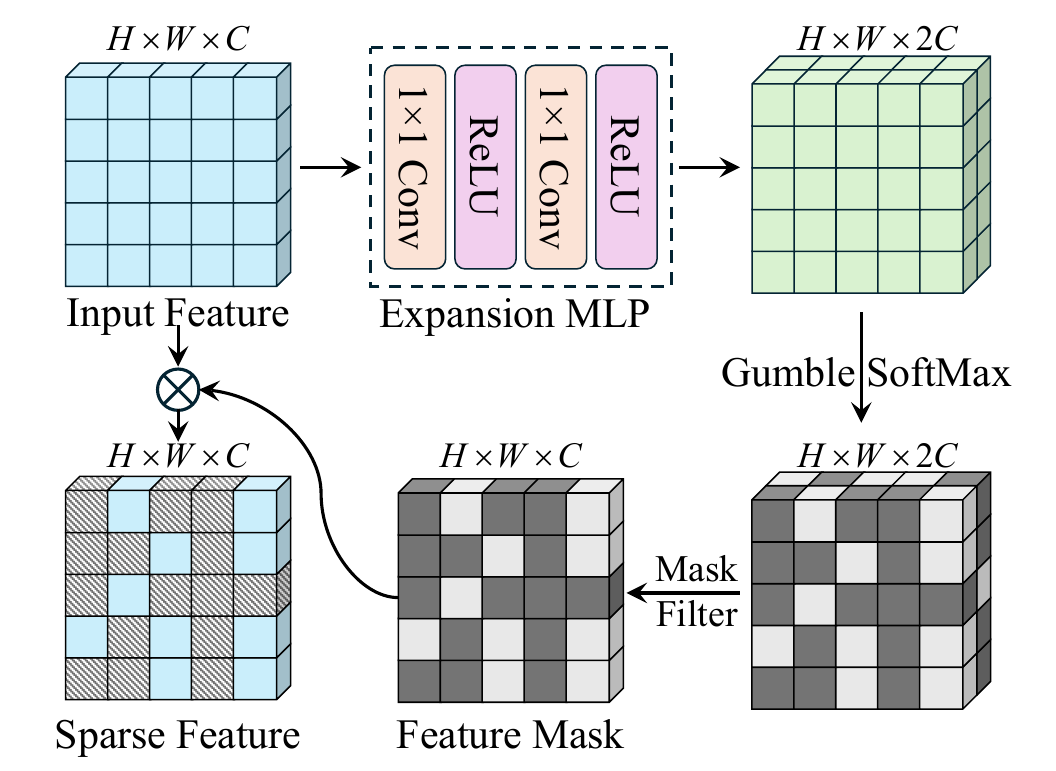}
    \caption{Architecture of the dynamic feature selector. A trainable MLP projects the input features to twice their original dimension. A pixel-wise Gumbel-Softmax operation is then applied to compute selection probabilities. }
    \label{fig:selector}
    \vspace{-3mm}
\end{wrapfigure}

%% file: figs/saliency.tex
\begin{wrapfigure}{r}{0.6\linewidth}
    \centering
    \vspace{-4mm}
    \includegraphics[width=1.0\linewidth]{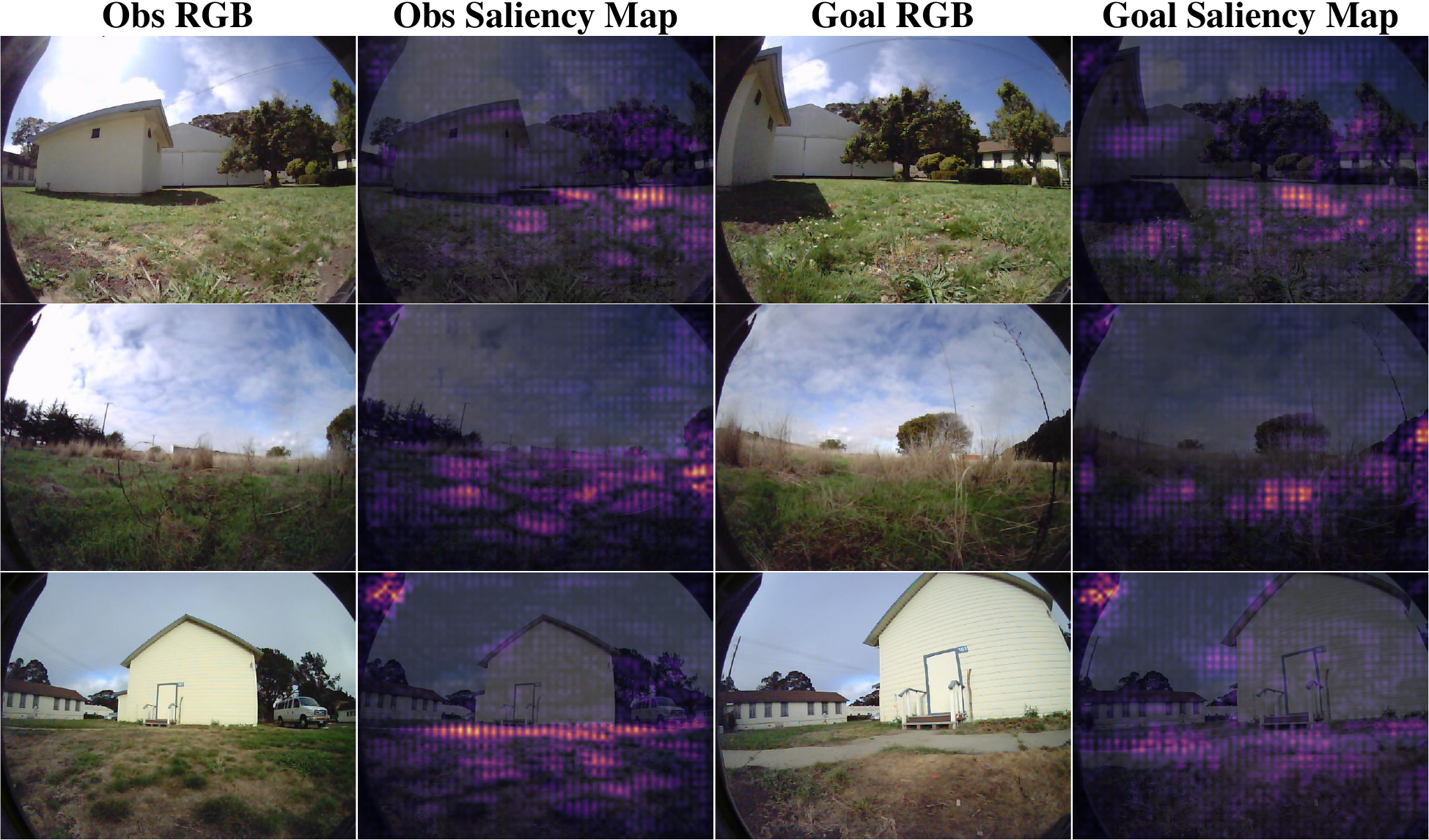}

    \caption{Visualization of saliency maps for observation and goal images.}
    \label{fig:saliency}
    % \vspace{-4mm}
\end{wrapfigure}

%% file: sec/4_experiment.tex
% \vspace{-3mm}
\section{Experiment}
\label{sec:experiment}
% \vspace{-3mm}
\subsection{Experimental Setups}
\subsubsection{Datasets}
% \vspace{-2mm}
We evaluated our method in two experimental settings: benchmark datasets and simulated environments. For the benchmark datasets, we select four diverse datasets to assess the performance of our approach under various conditions. These include the Recon dataset~\cite{shah2023vint}, which provides medium-speed {($2m/s$)} outdoor data to evaluate our method in real-world, dynamic outdoor settings, and the SCAND dataset~\cite{scand}, a medium-speed dataset featuring environmental interactions. Additionally, we include the Go-Stanford dataset~\cite{hirose2019deep} and the SACSoN dataset~\cite{hirose2023sacson}, representing low-speed {($0.5m/s$)} and medium-speed indoor scenarios, respectively. These datasets allow us to test our method across environments with varying speed characteristics. All datasets are pre-processed following the ViNT method~\cite{shah2023vint} to ensure consistency across experiments. For each dataset, we randomly split the data into training (80\%) and testing (20\%) sets. The implementation detail is illustrated in Section~\ref{sec:app_implement} in the Appendix.
% \vspace{-3mm}
\input{tables/quant}
\subsubsection{Evaluation Metrics}
% \vspace{-2mm}
% For the benchmark comparison, we report the cosine similarity of action angles and the cosine similarity of predicted waypoints then report as $\text{Sim}(\mathbf{a}_t, \mathbf{a}_t^{\text{gt}})$ and $ \text{Sim}(\mathbf{w}_t, \mathbf{w}_t^{\text{gt}})$, respectively. Since NoMad~\cite{sridhar2023nomad} only outputs waypoints through the diffusion process, we leave its action angle term blank. To emphasize our model's efficiency advantages, we report the FLOPs and training memory usage for each model. For inference, we record the time of predicting a single trajectory for each model. The loss values of the action vector and distance are also reported.
% For the CARLA~\cite{dosovitskiy2017carla} simulation, we record the progress of our model-driven agent until it reaches the target or encounters a collision. The success rate is the mean of dividing the progress length by the total trajectory length. 
For the benchmark comparison, we report the cosine similarity between action angles and predicted waypoints, denoted as $\text{Sim}(\mathbf{a}_t, \mathbf{a}_t^{\text{gt}})$ and $ \text{Sim}(\mathbf{w}_t, \mathbf{w}_t^{\text{gt}})$, respectively {(in percentage)}. Since NoMad~\cite{sridhar2023nomad} only outputs waypoints through the diffusion process, we omit the action angle term for this model. {To highlight the efficiency advantages of our approach, we report the FLOPs and memory usage of each model on the entire evaluation set.} For inference, we measure the time required to predict a single trajectory for each model. Additionally, we report the loss values for the action vector and distance.

For the CARLA~\cite{dosovitskiy2017carla} simulation, we track the progress of our model-driven agent until it either reaches the target or encounters a collision. The success rate is calculated as the mean of the ratio of progress length to total trajectory length.

% \vspace{-3mm}
\subsection{Evaluation in Real-world Benchmarks}
% \vspace{-2mm}
\input{figs/histcomp} Table~\ref{tb:quant} illustrates the performance of our method on RECON~\cite{shah2023vint}, Go-Stanford~\cite{shah2023gnm}, SACSoN~\cite{hirose2023sacson}, and SCAND~\cite{scand} datasets. We compare the performance with ViNT~\cite{shah2023vint} and NoMad~\cite{sridhar2023nomad}. All models are trained from scratch. Our model saves about 58\% FLOPs across all benchmarks compared to ViNT~\cite{shah2023vint} while maintaining comparable accuracy. Figure~\ref{fig:histcomp} depicts the efficiency advantages of our model compared to ViNT~\cite{shah2023vint}. Our approach achieves a 0.83\% improvement in $\text{Sim}(\mathbf{a}_t, \mathbf{a}_t^{\text{gt}})$ and a 0.28\% increase in $\text{Sim}(\mathbf{w}_t, \mathbf{w}_t^{\text{gt}})$ compared to ViNT~\cite{shah2023vint} across four benchmarks. In terms of time efficiency, we save 0.16 seconds compared to ViNT~\cite{shah2023vint} and 0.89 seconds compared to NoMaD~\cite{sridhar2023nomad}. Despite this, NoMaD~\cite{sridhar2023nomad}, due to its diffusion refinement procedure, achieves an average performance that is 0.2\% higher than ours. However, NoMaD~\cite{sridhar2023nomad} requires approximately four times the FLOPs of our method, making it less efficient. Notably, the average FLOPs of our dynamic model in RECON~\cite{shah2023vint} and SCAND~\cite{scand} are higher than those in SACSoN~\cite{hirose2023sacson} and Go-Stanford~\cite {shah2023gnm}. One reason for this is that the former two datasets are from outdoor environments, while the latter two consist of indoor scenarios. The indoor datasets benefit from lower speeds, more controllable environments, and less complex lighting conditions. This finding also validates the assumption that for a more complex scene, activating more layers for accurate navigation.

% \vspace{-4mm}
\subsection{Real-time Robotic Navigation in CARLA Simulation Environment}
% \vspace{-3mm}
\input{figs/carla_res_town02} For the simulation, we first collected 200 trajectories of inline navigation data from CARLA~\cite{dosovitskiy2017carla} Town01 to fine-tune the pre-trained model. The data was gathered using an RGB camera and various sensors mounted on an autopilot agent operating at a frequency of 4Hz. To ensure consistency, we standardized the image size to $640 \times 480$ pixels with a 90° field of view (FOV). We evaluated our method across three distinct CARLA environments: Town02 (Scene A), Town03 (Scene B), and Town10 (Scene C). Scene A, with a small-town layout and simple residential-commercial mix, represents an ``easy task" for the agent. Scene B, a larger urban map with roundabouts and large junctions, is considered a ``medium task." Scene C, a downtown area filled with skyscrapers, residential buildings, and parked cars, presents a highly dynamic and complex environment, making it a ``hard task." For each scene, we collected 20 trajectories, which were used in the subsequent testing phase. The car was driven by a BehaviorAgent~\cite{dosovitskiy2017carla}, maintaining a consistent maximum speed of 20 km/h across all environments. More details are illustrated in Appendix Section~\ref{sec:app_implement}.

Figure~\ref{fig:carla_res_town_02} illustrates the visualized navigation performance of baselines and ours. GNM~\cite{shah2023gnm} lacks enough generalization ability to achieve the target task. ViNT~\cite{shah2023vint} and our method can successfully reach the target points. Note that the trajectory of ViNT~\cite{shah2023vint} has some drift. This is due to ViNT~\cite{shah2023vint} utilizing all features and decoder layers, potentially overfitting to training data and producing suboptimal trajectories. Our approach dynamically activates transformer layers and selectively filters features, resulting in superior trajectory performance.

\input{tables/carla} Table~\ref{tb:carla} presents the success rate of different models on the CARLA~\cite{dosovitskiy2017carla} simulation. NoMad~\cite{sridhar2023nomad} is unable to achieve agile real-time simulation on our test platform due to the computationally intensive nature of its diffusion process. The results show that, although our model has higher FLOPs than GNM~\cite{shah2023gnm}, its success rate shows a 38\% improvement, demonstrating the effectiveness of our approach. Compared to ViNT~\cite{shah2023vint}, our method not only achieves comparable performance but also reduces FLOPs by more than a factor of two. Furthermore, as the simulation environment becomes more challenging (Scene A$\rightarrow$Scene C), the FLOPs required by our model increase. This is because we use a unified early exit metric across all three simulation environments. As the visual discrepancy between the observation and goal increases, our model needs more decoder blocks to extract contextual information effectively.

% \vspace{-3mm}
\subsection{Ablation Study}
% \vspace{-2mm}
\textbf{Ablation into Individual Modules:}
Table~\ref{tb:ablation_arch} illustrates the effectiveness of our proposed module. The dynamic decoder column represents whether we are using early exit on the transformer decoder. {The first row shows the result when we simply deactivate half of the decoder layers. Similarly, the second row presents the result when we deliberately reduce the hidden channel size from $C$ to $\frac{C}{2}$. Although these settings can improve efficiency, they usually lead to decreased performance and poor generalization (i.e., high accuracy on the training set but low accuracy on the testing set).}. The rest of Table~\ref{tb:ablation_arch} elaborates that without the dynamic decoder, the efficiency does not vary too much compared to the baseline. Moreover, by using the feature selector, the performance will be better, and the efficiency will also be boosted. This is because our proposed feature selector sparsifies the features and stabilizes the early exit process. 
\input{tables/ablation_arch}

\input{tables/ablation_extra} \textbf{Ablation into Threshold Optimization:} Table~\ref{tb:ablation_extra} shows the different performances of whether we implement an extra Bayesian Optimization (BO) after training as DeeR-VLA~\cite{DeeR-VLA}. Moreover, it reports the influence of whether we allow an early exit before the decoder. Results show that without BO, the early exit process can be impaired due to a suboptimal threshold. Besides, if we allow the early exit before the transformer decoder, although it can gain efficiency improvement, the overall accuracy will slightly decrease. Therefore, such a technique ought to be a trade-off that requires careful design.

\begin{figure}
    \centering
    \includegraphics[width=1.0\linewidth]{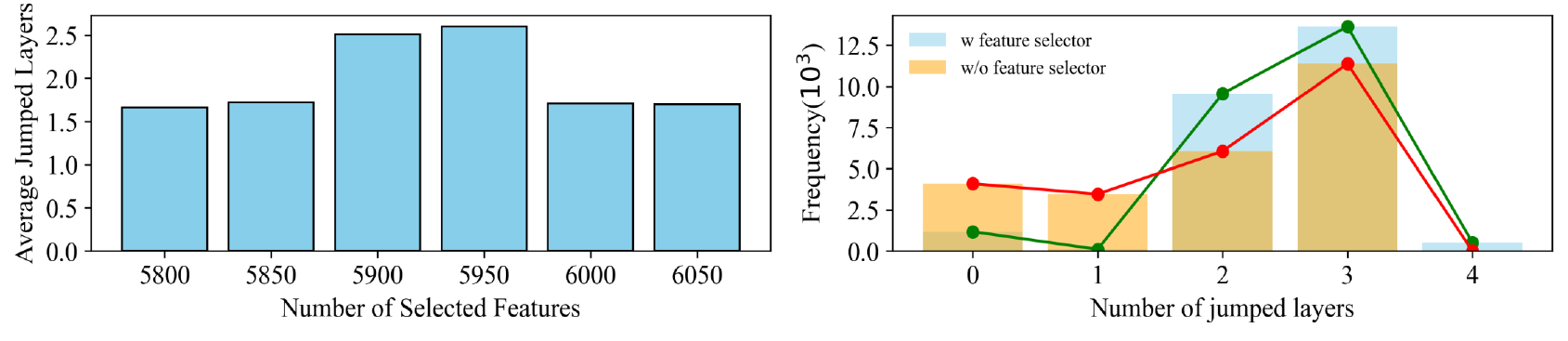}
    \caption{Visualized relationship between the number of selected features and the skipped layers (\textbf{Left}) and the frequency of different numbers of skipped layers in the context of with or without the feature selector (\textbf{right}).}
    \label{fig:jumpedlayers}
    \label{fig:masknum}
    % \vspace{-5.5mm}
\end{figure}

\textbf{Ablation into Feature Selection for Early Exit:}
To assess the impact of our proposed feature selector on the early exit mechanism, we evaluate the model on the RECON~\cite{shah2023vint} test set with a batch size of 1, both with and without the feature selector. For each sample, we record the MEAN value of the action difference between each layer (0-1,1-2,2-3). Moreover, we record the early exit index, referring to it as the number of skipped layers. {In Figure~\ref{fig:masknum}, we observe that within a certain range of selected feature numbers, the average number of skipped layers remains high. This finding suggests that our feature selector helps determine an appropriate early exit threshold, thereby enhancing the frequency of early exiting.} Additionally, Figure~\ref{fig:jumpedlayers} shows that our proposed feature selector increases the frequency of 2-to-4 layer jumps, leading to improved efficiency. Therefore, by integrating the feature selector with early exit, our method achieves both more stable and more efficient performance.

% \begin{figure}[tbp]
%   \centering
%   \begin{minipage}{0.47\linewidth}
%     \centering
%     \includegraphics[width=1.0\linewidth]{figs/numvslayer.png}
%     \caption{\textcolor{red}{Visualized relationship between the number of selected features and the skipped layers. We scatter all samples, then bin them in steps of 50.} }
%     \label{fig:masknum}
%   \end{minipage}
%   \hfill
%   \begin{minipage}{0.52\linewidth}
%     \centering
%     \includegraphics[width=1.0\linewidth]{figs/jumpedlayers2.png}
%     \caption{Visualization of the frequency of different numbers of skipped layers in the context of with or without the feature selector.}
%     \label{fig:jumpedlayers}
%   \end{minipage}
% \end{figure}

%% file: tables/quant.tex
\begin{table*}[!t]
      \caption{Quantitative Comparison on Benchmarks.  We highlight our method with the \colorbox{iccvblue!15}{colored font}, the best and the second best value of each metric are reported with \textbf{bold} and \underline{underlined} fonts, respectively.}
    \resizebox{1\textwidth}{!}{

     \begin{tabular}{c c c c c c c c c}
    \toprule
    Dataset&Method& $\text{Sim}(\mathbf{a}_t, \mathbf{a}_t^{\text{gt}}) \uparrow$&$ \text{Sim}(\mathbf{w}_t, \mathbf{w}_t^{\text{gt}}) \uparrow$ &$\mathcal{L}_{action}$$\downarrow$& $\mathcal{L}_{dist}$$\downarrow$ &FLOPs ($10^9$) &Time(s/traj) &Memory (Gb)\\
    \midrule
        % &GNM&90.55&95.13&0.238&12.96&1.14\\
    \multirow{3}{*}{Recon~\cite{shah2023vint}}
    &ViNT~\cite{shah2023vint}&\underline{94.49}&96.20&0.285&6.94&\underline{4.37}&\underline{0.379}&\underline{19.07}\\
    &NoMad~\cite{sridhar2023nomad}&- &\textbf{96.64}&\underline{0.207}&\underline{6.44}&7.46 &1.118 &21.36\\
    \rowcolor{iccvblue!15}\cellcolor[RGB]{255,255,255}&\textbf{Ours}& \textbf{{94.92}}&\underline{96.53}&\textbf{0.191}&\textbf{6.26}&\textbf{1.93}&\textbf{0.228}&\textbf{13.35}\\
    \midrule

    % &GNM&84.70&94.00&0.514&18.51&1.14&&\\
    \multirow{3}{*}{Go-Stanford~\cite{shah2023gnm}}&ViNT~\cite{shah2023vint}&\underline{88.50}&93.47&0.531&15.80&\underline{4.37}&\underline{0.379}&\underline{19.07}\\
    &NoMad~\cite{sridhar2023nomad}&-&\underline{93.51}&\underline{0.507} &\textbf{12.93}&7.46&1.118&21.36\\
    \rowcolor{iccvblue!15}\cellcolor[RGB]{255,255,255}&\textbf{Ours}& \textbf{89.07}& \textbf{93.66}&\textbf{0.449}&\underline{14.23}&\textbf{1.68}&\textbf{0.209} &\textbf{12.27}\\

    \midrule
    
    % &GNM&88.70&92.98&0.755&10.31&1.14&&\\
    \multirow{3}{*}{SACSoN~\cite{hirose2023sacson}}&ViNT~\cite{shah2023vint}&89.66&93.16&0.686&10.95&\underline{4.37}&\underline{0.379}&\underline{19.07}\\
    &NoMad~\cite{sridhar2023nomad}&-&\underline{93.69}&\underline{0.501} &\underline{9.66}&7.46&1.118&21.36\\
    \rowcolor{iccvblue!15}\cellcolor[RGB]{255,255,255}&   \textbf{Ours}& \textbf{{90.54}}& \textbf{93.72}&\textbf{0.493}&\textbf{9.62}&\textbf{1.68}&\textbf{0.209} &\textbf{12.27}\\
    \midrule

    % &GNM&94.18&97.25&0.187&15.54&1.14\\
    \multirow{3}{*}{SCAND~\cite{scand}}&ViNT~\cite{shah2023vint}&\underline{95.43}&96.89&\underline{0.141}&16.08&\underline{4.37}&\underline{0.379}&\underline{19.07}\\
    &NoMad~\cite{sridhar2023nomad}&-&\textbf{97.79}&\textbf{0.121} &\textbf{13.05}&7.46&1.118&21.36\\
    \rowcolor{iccvblue!15}\cellcolor[RGB]{255,255,255}& \textbf{Ours}& \textbf{96.85}& \underline{97.03}&\underline{0.130}& \underline{14.41}& \textbf{1.93}&\textbf{0.228}&\textbf{13.25}\\
    
    \bottomrule
    
  \end{tabular}}

  \label{tb:quant}
  % \vspace{-5mm}
\end{table*}

%% file: figs/histcomp.tex
\begin{wrapfigure}{r}{0.55\linewidth}
    \centering
    \vspace{-3mm}
    \includegraphics[width=1.0\linewidth]{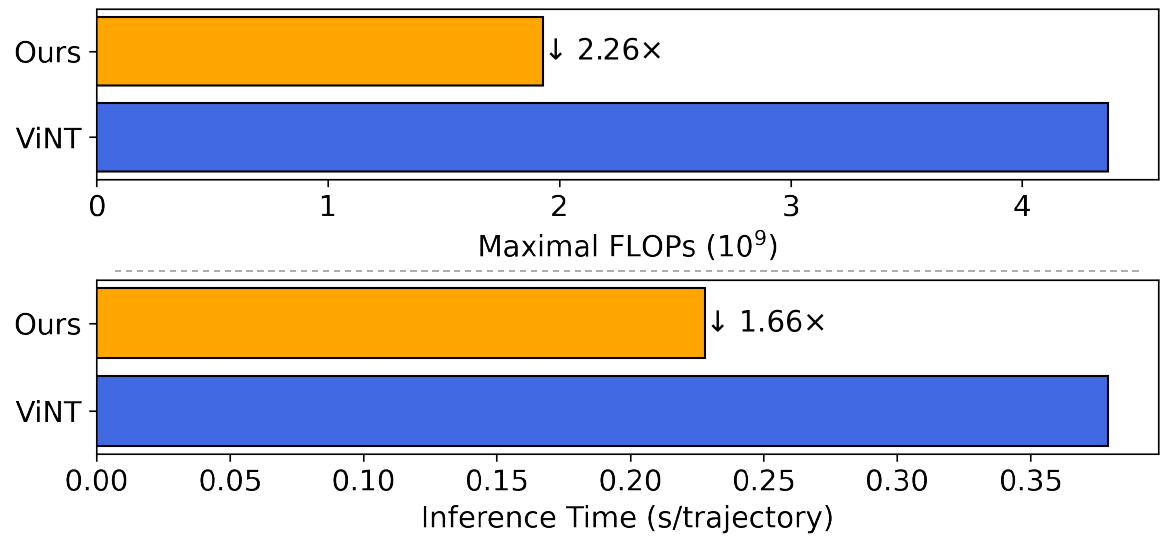}

    \caption{Efficiency comparison of \textbf{upper:} FLOPs, \textbf{bottom:} inference time between ours and ViNT on RECON dataset.}
    \label{fig:histcomp}
    \vspace{-3mm}
\end{wrapfigure}

%% file: figs/carla_res_town02.tex
\begin{wrapfigure}{r}{0.6\linewidth}
    \centering
    \vspace{-4mm}
    \includegraphics[width=1.0\linewidth]{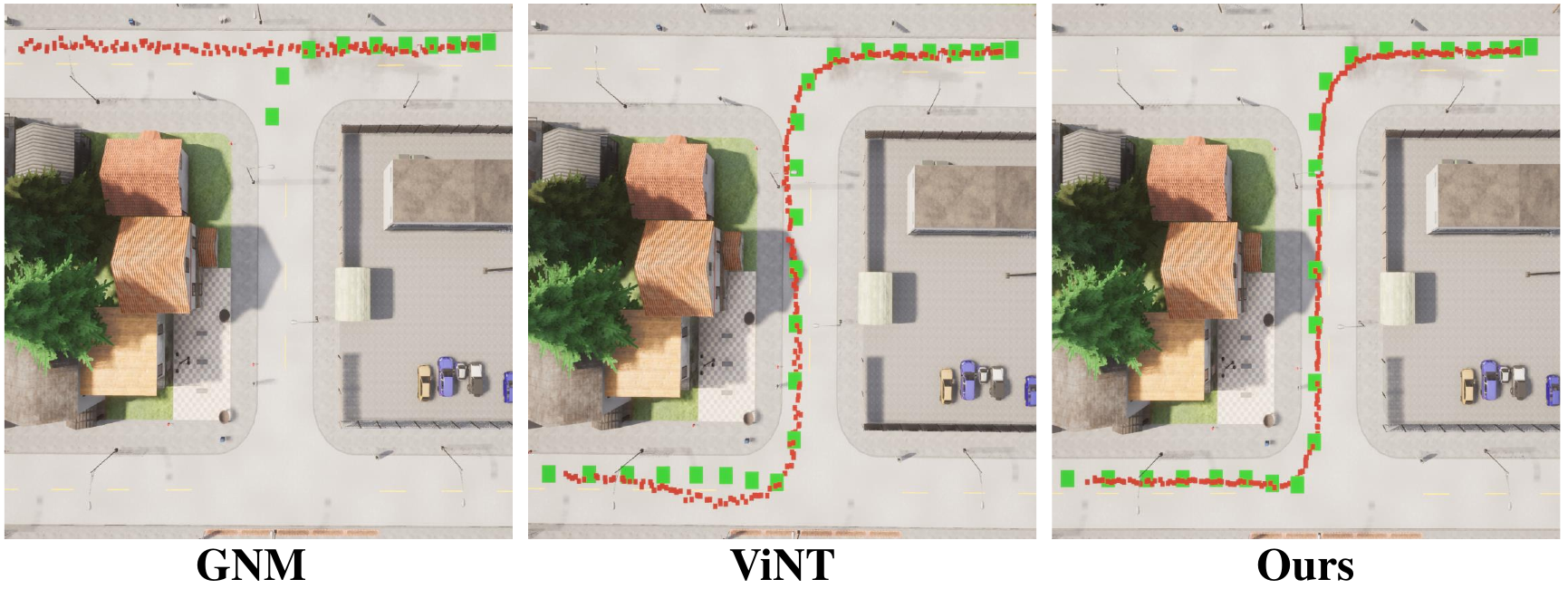}
    
    \caption{The simulation result in CARLA Town02 environment. The green dots represent the discrete goals, and the red dots represent the predicted waypoints.}
    \label{fig:carla_res_town_02}
    % \vspace{-4mm}
\end{wrapfigure}

%% file: tables/carla.tex
\begin{wraptable}{r}{0.55\linewidth}{
\centering 
\vspace{-4mm}
\caption{The comparison of our model with baselines in the CARLA under various environments. The best and the second best values of each metric are reported with \textbf{bold} and \underline{underlined} fonts, respectively.}
\resizebox{0.55\textwidth}{!}{
     \begin{tabular}{c c c c c c c c c}
    \toprule
    Environment &Model& Successful Rate & FLOPs ($10^9$)\\
    \midrule
    \multirow{3}{*}{Scene A}&GNM~\cite{shah2023gnm} & 0.297 & 1.09\\
    &ViNT~\cite{shah2023vint} & \underline{0.724} & 4.37\\
    \rowcolor{iccvblue!15}\cellcolor[RGB]{255,255,255}&Ours &\textbf{0.727} &1.58\\
    \midrule
    
    \multirow{3}{*}{Scene B}&GNM~\cite{shah2023gnm} & 0.288 & 1.09\\
    &ViNT~\cite{shah2023vint} & \underline{0.659} & 4.37\\
    \rowcolor{iccvblue!15}\cellcolor[RGB]{255,255,255}&Ours & \textbf{0.664} &1.70\\
    \midrule
    
    \multirow{3}{*}{Scene C}&GNM~\cite{shah2023gnm} & 0.251 & 1.09\\
    &ViNT~\cite{shah2023vint} & \textbf{0.589} & 4.37\\
    \rowcolor{iccvblue!15}\cellcolor[RGB]{255,255,255}&Ours &\underline{0.588} &1.93 \\
    \bottomrule
  \end{tabular}}
  \label{tb:carla}}
  % \vspace{-3mm}
\end{wraptable}

%% file: tables/ablation_arch.tex
\begin{table*}[!t]
\centering

    \caption{Ablation study of the effectiveness of individual modules on the RECON dataset.}

\resizebox{1\textwidth}{!}{
     \begin{tabular}{c c c c c c c c c c}
    \toprule
    {Dynamic decoder}& {Feature selector}& $\text{Sim}(\mathbf{a}_t, \mathbf{a}_t^{\text{gt}})$&$ \text{Sim}(\mathbf{w}_t, \mathbf{w}_t^{\text{gt}})$ &$\mathcal{L}_{action}$& $\mathcal{L}_{dist}$ &FLOPs ($10^9$) &Time &Memory\\
    \midrule
    {Half layers} & - & 91.05& 93.28 &0.332&7.53&{2.61}&{0.306}&{17.48}\\
    {Half channel} & - & 89.70 & 92.41 &0.390&7.71&{2.19}&{0.270}&{12.11}\\
    -&-&94.49&96.20&0.285&6.94&{4.37}&{0.379}&{19.07}\\
    \checkmark&-&93.68 &95.42 &0.274 &7.08 &2.41 &0.251 &16.49\\
    -&\checkmark&94.81 &96.44 &0.205 &6.30 &4.06 &0.377 &18.22\\
    \rowcolor{iccvblue!15}\checkmark&\checkmark&\textbf{94.92} &\textbf{96.53} &\textbf{0.191} &\textbf{6.26} &\textbf{1.93} &\textbf{0.228} &\textbf{13.35}\\
    \bottomrule
  \end{tabular}}
  \label{tb:ablation_arch}
  % \vspace{-3mm}
\end{table*}

%% file: tables/ablation_extra.tex
\begin{wraptable}{r}{0.6\linewidth}
\centering
\vspace{-4mm}
\caption{Ablation study on whether using post-training Bayesian Optimization (BO) and allowing exit before the decoder. The best and the second best values of each metric are reported with \textbf{bold} and \underline{underlined} fonts, respectively.}
\resizebox{1.0\linewidth}{!}{
     \begin{tabular}{c c c c c c c c c}
    \toprule
    BO & Pre-decoder Exit &$ \text{Sim}(\mathbf{w}_t, \mathbf{w}_t^{\text{gt}})$ & Successful Rate & FLOPs ($10^9$)\\
    \midrule
        - & - & 96.30 & 0.725 &2.46\\
        -& \checkmark& 96.22 & 0.719 & 2.27\\
        \checkmark& -&\textbf{96.58} & \textbf{0.732} & \underline{2.11}\\
        \checkmark& \checkmark& \underline{96.53 } & \underline{0.727} & \textbf{1.93}\\
    \bottomrule
  \end{tabular}}
  \label{tb:ablation_extra}
  % \vspace{-4mm}
\end{wraptable}

%% file: sec/5_conclusion.tex
% \vspace{-4.5mm}
\section{Conclusion}\label{sec:conclusion}
% \vspace{-3.5mm}
In this work, we propose DynaNav, a novel, highly efficient visual navigation model. We first introduce a dynamic feature selector that filters observations and goals to extract robust, memory-efficient features. We also introduce feature-aware early exit criteria for the transformer decoders, using action consistency metrics optimized via Bayesian techniques. Our experimental results show a significant reduction in computational overhead compared to existing foundation navigation models while maintaining high performance across standard benchmarks and the CARLA simulation environment. The empirical evidence validates the effectiveness of our approach in achieving efficient and robust visual navigation.

To achieve optimal performance, our model requires an additional optimization process. Although the Bayesian optimization helps fine-tune the model and determine optimal thresholds, the added labor cost is non-negligible. Future work could involve implementing these optimization techniques concurrently with training to create a more streamlined end-to-end system. Furthermore, the proposed feature selection mechanism can be integrated with various CNN-based encoder models to improve their overall efficiency.

%% file: sec/6_acknowledgement.tex
\section{Acknowledgement}
This research is supported by the National University of Singapore under the NUS College of Design and Engineering Industry-focused Ring-Fenced PhD Scholarship programme. Changhao Chen is funded by the Young Elite Scientist Sponsorship Program by CAST (No. YESS20220181) and the National Natural Science Foundation of China (NFSC) under the Grant Number 62573370.

%% file: sec/checklist.tex
\section*{NeurIPS Paper Checklist}

\begin{enumerate}

\item {\bf Claims}
    \item[] Question: Do the main claims made in the abstract and introduction accurately reflect the paper's contributions and scope?
    \item[] Answer: \answerYes{} % Replace by \answerYes{}, \answerNo{}, or \answerNA{}.
    \item[] Justification: We develop a novel method for efficient visual navigation. Our proposed method effectively reduces the computational and time cost by using the developed feature selector to obtain sparse features and utilizing dynamic early-exiting to skip decoder layers.
    \item[] Guidelines:
    \begin{itemize}
        \item The answer NA means that the abstract and introduction do not include the claims made in the paper.
        \item The abstract and/or introduction should clearly state the claims made, including the contributions made in the paper and important assumptions and limitations. A No or NA answer to this question will not be perceived well by the reviewers. 
        \item The claims made should match theoretical and experimental results, and reflect how much the results can be expected to generalize to other settings. 
        \item It is fine to include aspirational goals as motivation as long as it is clear that these goals are not attained by the paper. 
    \end{itemize}

\item {\bf Limitations}
    \item[] Question: Does the paper discuss the limitations of the work performed by the authors?
    \item[] Answer: \answerYes{} % Replace by \answerYes{}, \answerNo{}, or \answerNA{}.
    \item[] Justification: In the conclusion section of the main body, we elaborate on some limitations of our work. These limitations can be further investigated to find suitable solutions.
    \item[] Guidelines:
    \begin{itemize}
        \item The answer NA means that the paper has no limitation while the answer No means that the paper has limitations, but those are not discussed in the paper. 
        \item The authors are encouraged to create a separate "Limitations" section in their paper.
        \item The paper should point out any strong assumptions and how robust the results are to violations of these assumptions (e.g., independence assumptions, noiseless settings, model well-specification, asymptotic approximations only holding locally). The authors should reflect on how these assumptions might be violated in practice and what the implications would be.
        \item The authors should reflect on the scope of the claims made, e.g., if the approach was only tested on a few datasets or with a few runs. In general, empirical results often depend on implicit assumptions, which should be articulated.
        \item The authors should reflect on the factors that influence the performance of the approach. For example, a facial recognition algorithm may perform poorly when image resolution is low or images are taken in low lighting. Or a speech-to-text system might not be used reliably to provide closed captions for online lectures because it fails to handle technical jargon.
        \item The authors should discuss the computational efficiency of the proposed algorithms and how they scale with dataset size.
        \item If applicable, the authors should discuss possible limitations of their approach to address problems of privacy and fairness.
        \item While the authors might fear that complete honesty about limitations might be used by reviewers as grounds for rejection, a worse outcome might be that reviewers discover limitations that aren't acknowledged in the paper. The authors should use their best judgment and recognize that individual actions in favor of transparency play an important role in developing norms that preserve the integrity of the community. Reviewers will be specifically instructed to not penalize honesty concerning limitations.
    \end{itemize}

\item {\bf Theory assumptions and proofs}
    \item[] Question: For each theoretical result, does the paper provide the full set of assumptions and a complete (and correct) proof?
    \item[] Answer: \answerNA{} % Replace by \answerYes{}, \answerNo{}, or \answerNA{}.
    \item[] Justification: In this work, we develop a novel architecture of the visual navigation model. There are no additional newly proposed theories, such as optimization or representation, etc.
    \item[] Guidelines:
    \begin{itemize}
        \item The answer NA means that the paper does not include theoretical results. 
        \item All the theorems, formulas, and proofs in the paper should be numbered and cross-referenced.
        \item All assumptions should be clearly stated or referenced in the statement of any theorems.
        \item The proofs can either appear in the main paper or the supplemental material, but if they appear in the supplemental material, the authors are encouraged to provide a short proof sketch to provide intuition. 
        \item Inversely, any informal proof provided in the core of the paper should be complemented by formal proofs provided in appendix or supplemental material.
        \item Theorems and Lemmas that the proof relies upon should be properly referenced. 
    \end{itemize}

    \item {\bf Experimental result reproducibility}
    \item[] Question: Does the paper fully disclose all the information needed to reproduce the main experimental results of the paper to the extent that it affects the main claims and/or conclusions of the paper (regardless of whether the code and data are provided or not)?
    \item[] Answer: \answerYes{} % Replace by \answerYes{}, \answerNo{}, or \answerNA{}.
    \item[] Justification: In the experiment section and appendix, we provide detailed settings for training and data processing.
    \item[] Guidelines:
    \begin{itemize}
        \item The answer NA means that the paper does not include experiments.
        \item If the paper includes experiments, a No answer to this question will not be perceived well by the reviewers: Making the paper reproducible is important, regardless of whether the code and data are provided or not.
        \item If the contribution is a dataset and/or model, the authors should describe the steps taken to make their results reproducible or verifiable. 
        \item Depending on the contribution, reproducibility can be accomplished in various ways. For example, if the contribution is a novel architecture, describing the architecture fully might suffice, or if the contribution is a specific model and empirical evaluation, it may be necessary to either make it possible for others to replicate the model with the same dataset, or provide access to the model. In general. releasing code and data is often one good way to accomplish this, but reproducibility can also be provided via detailed instructions for how to replicate the results, access to a hosted model (e.g., in the case of a large language model), releasing of a model checkpoint, or other means that are appropriate to the research performed.
        \item While NeurIPS does not require releasing code, the conference does require all submissions to provide some reasonable avenue for reproducibility, which may depend on the nature of the contribution. For example
        \begin{enumerate}
            \item If the contribution is primarily a new algorithm, the paper should make it clear how to reproduce that algorithm.
            \item If the contribution is primarily a new model architecture, the paper should describe the architecture clearly and fully.
            \item If the contribution is a new model (e.g., a large language model), then there should either be a way to access this model for reproducing the results or a way to reproduce the model (e.g., with an open-source dataset or instructions for how to construct the dataset).
            \item We recognize that reproducibility may be tricky in some cases, in which case authors are welcome to describe the particular way they provide for reproducibility. In the case of closed-source models, it may be that access to the model is limited in some way (e.g., to registered users), but it should be possible for other researchers to have some path to reproducing or verifying the results.
        \end{enumerate}
    \end{itemize}

\item {\bf Open access to data and code}
    \item[] Question: Does the paper provide open access to the data and code, with sufficient instructions to faithfully reproduce the main experimental results, as described in supplemental material?
    \item[] Answer: \answerNo{} % Replace by \answerYes{}, \answerNo{}, or \answerNA{}.
    \item[] Justification: All data we use in this paper comes from publicly available datasets. Upon the situation of acceptance, we will consider releasing the code.
    \item[] Guidelines:
    \begin{itemize}
        \item The answer NA means that paper does not include experiments requiring code.
        \item Please see the NeurIPS code and data submission guidelines (\url{https://nips.cc/public/guides/CodeSubmissionPolicy}) for more details.
        \item While we encourage the release of code and data, we understand that this might not be possible, so “No” is an acceptable answer. Papers cannot be rejected simply for not including code, unless this is central to the contribution (e.g., for a new open-source benchmark).
        \item The instructions should contain the exact command and environment needed to run to reproduce the results. See the NeurIPS code and data submission guidelines (\url{https://nips.cc/public/guides/CodeSubmissionPolicy}) for more details.
        \item The authors should provide instructions on data access and preparation, including how to access the raw data, preprocessed data, intermediate data, and generated data, etc.
        \item The authors should provide scripts to reproduce all experimental results for the new proposed method and baselines. If only a subset of experiments are reproducible, they should state which ones are omitted from the script and why.
        \item At submission time, to preserve anonymity, the authors should release anonymized versions (if applicable).
        \item Providing as much information as possible in supplemental material (appended to the paper) is recommended, but including URLs to data and code is permitted.
    \end{itemize}

\item {\bf Experimental setting/details}
    \item[] Question: Does the paper specify all the training and test details (e.g., data splits, hyperparameters, how they were chosen, type of optimizer, etc.) necessary to understand the results?
    \item[] Answer: \answerYes{} % Replace by \answerYes{}, \answerNo{}, or \answerNA{}.
    \item[] Justification: In the experiment section and appendix, we provide detailed settings for training and data processing.
    \item[] Guidelines:
    \begin{itemize}
        \item The answer NA means that the paper does not include experiments.
        \item The experimental setting should be presented in the core of the paper to a level of detail that is necessary to appreciate the results and make sense of them.
        \item The full details can be provided either with the code, in appendix, or as supplemental material.
    \end{itemize}

\item {\bf Experiment statistical significance}
    \item[] Question: Does the paper report error bars suitably and correctly defined or other appropriate information about the statistical significance of the experiments?
    \item[] Answer: \answerYes{} % Replace by \answerYes{}, \answerNo{}, or \answerNA{}.
    \item[] Justification: We use the mean value as the justification metric. For each dataset, the final metric value is the mean across all samples. As the parameters of the model are frozen during the inference, therefore, the standard deviation of the same model inference on the same dataset is small enough.
    \item[] Guidelines:
    \begin{itemize}
        \item The answer NA means that the paper does not include experiments.
        \item The authors should answer "Yes" if the results are accompanied by error bars, confidence intervals, or statistical significance tests, at least for the experiments that support the main claims of the paper.
        \item The factors of variability that the error bars are capturing should be clearly stated (for example, train/test split, initialization, random drawing of some parameter, or overall run with given experimental conditions).
        \item The method for calculating the error bars should be explained (closed form formula, call to a library function, bootstrap, etc.)
        \item The assumptions made should be given (e.g., Normally distributed errors).
        \item It should be clear whether the error bar is the standard deviation or the standard error of the mean.
        \item It is OK to report 1-sigma error bars, but one should state it. The authors should preferably report a 2-sigma error bar than state that they have a 96\% CI, if the hypothesis of Normality of errors is not verified.
        \item For asymmetric distributions, the authors should be careful not to show in tables or figures symmetric error bars that would yield results that are out of range (e.g. negative error rates).
        \item If error bars are reported in tables or plots, The authors should explain in the text how they were calculated and reference the corresponding figures or tables in the text.
    \end{itemize}

\item {\bf Experiments compute resources}
    \item[] Question: For each experiment, does the paper provide sufficient information on the computer resources (type of compute workers, memory, time of execution) needed to reproduce the experiments?
    \item[] Answer: \answerYes{} % Replace by \answerYes{}, \answerNo{}, or \answerNA{}.
    \item[] Justification: We illustrate all hardware resources we used in the experimental section and appendix.
    \item[] Guidelines:
    \begin{itemize}
        \item The answer NA means that the paper does not include experiments.
        \item The paper should indicate the type of compute workers CPU or GPU, internal cluster, or cloud provider, including relevant memory and storage.
        \item The paper should provide the amount of compute required for each of the individual experimental runs as well as estimate the total compute. 
        \item The paper should disclose whether the full research project required more compute than the experiments reported in the paper (e.g., preliminary or failed experiments that didn't make it into the paper). 
    \end{itemize}
    
\item {\bf Code of ethics}
    \item[] Question: Does the research conducted in the paper conform, in every respect, with the NeurIPS Code of Ethics \url{https://neurips.cc/public/EthicsGuidelines}?
    \item[] Answer: \answerYes{} % Replace by \answerYes{}, \answerNo{}, or \answerNA{}.
    \item[] Justification: Our work conforms to it in every respect. 
    \item[] Guidelines:
    \begin{itemize}
        \item The answer NA means that the authors have not reviewed the NeurIPS Code of Ethics.
        \item If the authors answer No, they should explain the special circumstances that require a deviation from the Code of Ethics.
        \item The authors should make sure to preserve anonymity (e.g., if there is a special consideration due to laws or regulations in their jurisdiction).
    \end{itemize}

\item {\bf Broader impacts}
    \item[] Question: Does the paper discuss both potential positive societal impacts and negative societal impacts of the work performed?
    \item[] Answer: \answerNA{} % Replace by \answerYes{}, \answerNo{}, or \answerNA{}.
    \item[] Justification: Our work is a standard vision task and does not involve societal issues.
    \item[] Guidelines:
    \begin{itemize}
        \item The answer NA means that there is no societal impact of the work performed.
        \item If the authors answer NA or No, they should explain why their work has no societal impact or why the paper does not address societal impact.
        \item Examples of negative societal impacts include potential malicious or unintended uses (e.g., disinformation, generating fake profiles, surveillance), fairness considerations (e.g., deployment of technologies that could make decisions that unfairly impact specific groups), privacy considerations, and security considerations.
        \item The conference expects that many papers will be foundational research and not tied to particular applications, let alone deployments. However, if there is a direct path to any negative applications, the authors should point it out. For example, it is legitimate to point out that an improvement in the quality of generative models could be used to generate deepfakes for disinformation. On the other hand, it is not needed to point out that a generic algorithm for optimizing neural networks could enable people to train models that generate Deepfakes faster.
        \item The authors should consider possible harms that could arise when the technology is being used as intended and functioning correctly, harms that could arise when the technology is being used as intended but gives incorrect results, and harms following from (intentional or unintentional) misuse of the technology.
        \item If there are negative societal impacts, the authors could also discuss possible mitigation strategies (e.g., gated release of models, providing defenses in addition to attacks, mechanisms for monitoring misuse, mechanisms to monitor how a system learns from feedback over time, improving the efficiency and accessibility of ML).
    \end{itemize}
    
\item {\bf Safeguards}
    \item[] Question: Does the paper describe safeguards that have been put in place for responsible release of data or models that have a high risk for misuse (e.g., pretrained language models, image generators, or scraped datasets)?
    \item[] Answer: \answerNA{} % Replace by \answerYes{}, \answerNo{}, or \answerNA{}.
    \item[] Justification: All data and models used by this work are publicly available and tested in many applications. There are no such risks for this paper.
    \item[] Guidelines:
    \begin{itemize}
        \item The answer NA means that the paper poses no such risks.
        \item Released models that have a high risk for misuse or dual-use should be released with necessary safeguards to allow for controlled use of the model, for example by requiring that users adhere to usage guidelines or restrictions to access the model or implementing safety filters. 
        \item Datasets that have been scraped from the Internet could pose safety risks. The authors should describe how they avoided releasing unsafe images.
        \item We recognize that providing effective safeguards is challenging, and many papers do not require this, but we encourage authors to take this into account and make a best faith effort.
    \end{itemize}

\item {\bf Licenses for existing assets}
    \item[] Question: Are the creators or original owners of assets (e.g., code, data, models), used in the paper, properly credited and are the license and terms of use explicitly mentioned and properly respected?
    \item[] Answer: \answerYes{} % Replace by \answerYes{}, \answerNo{}, or \answerNA{}.
    \item[] Justification: We use publicly available code resources.
    \item[] Guidelines:
    \begin{itemize}
        \item The answer NA means that the paper does not use existing assets.
        \item The authors should cite the original paper that produced the code package or dataset.
        \item The authors should state which version of the asset is used and, if possible, include a URL.
        \item The name of the license (e.g., CC-BY 4.0) should be included for each asset.
        \item For scraped data from a particular source (e.g., website), the copyright and terms of service of that source should be provided.
        \item If assets are released, the license, copyright information, and terms of use in the package should be provided. For popular datasets, \url{paperswithcode.com/datasets} has curated licenses for some datasets. Their licensing guide can help determine the license of a dataset.
        \item For existing datasets that are re-packaged, both the original license and the license of the derived asset (if it has changed) should be provided.
        \item If this information is not available online, the authors are encouraged to reach out to the asset's creators.
    \end{itemize}

\item {\bf New assets}
    \item[] Question: Are new assets introduced in the paper well documented and is the documentation provided alongside the assets?
    \item[] Answer: \answerNA{} % Replace by \answerYes{}, \answerNo{}, or \answerNA{}.
    \item[] Justification: The paper does not release new assets.
    \item[] Guidelines:
    \begin{itemize}
        \item The answer NA means that the paper does not release new assets.
        \item Researchers should communicate the details of the dataset/code/model as part of their submissions via structured templates. This includes details about training, license, limitations, etc. 
        \item The paper should discuss whether and how consent was obtained from people whose asset is used.
        \item At submission time, remember to anonymize your assets (if applicable). You can either create an anonymized URL or include an anonymized zip file.
    \end{itemize}

\item {\bf Crowdsourcing and research with human subjects}
    \item[] Question: For crowdsourcing experiments and research with human subjects, does the paper include the full text of instructions given to participants and screenshots, if applicable, as well as details about compensation (if any)? 
    \item[] Answer: \answerNA{} % Replace by \answerYes{}, \answerNo{}, or \answerNA{}.
    \item[] Justification:  The paper does not involve crowdsourcing nor research with human subjects
    \item[] Guidelines:
    \begin{itemize}
        \item The answer NA means that the paper does not involve crowdsourcing nor research with human subjects.
        \item Including this information in the supplemental material is fine, but if the main contribution of the paper involves human subjects, then as much detail as possible should be included in the main paper. 
        \item According to the NeurIPS Code of Ethics, workers involved in data collection, curation, or other labor should be paid at least the minimum wage in the country of the data collector. 
    \end{itemize}

\item {\bf Institutional review board (IRB) approvals or equivalent for research with human subjects}
    \item[] Question: Does the paper describe potential risks incurred by study participants, whether such risks were disclosed to the subjects, and whether Institutional Review Board (IRB) approvals (or an equivalent approval/review based on the requirements of your country or institution) were obtained?
    \item[] Answer: \answerNA{} % Replace by \answerYes{}, \answerNo{}, or \answerNA{}.
    \item[] Justification: The paper does not involve crowdsourcing nor research with human subjects.
    \item[] Guidelines:
    \begin{itemize}
        \item The answer NA means that the paper does not involve crowdsourcing nor research with human subjects.
        \item Depending on the country in which research is conducted, IRB approval (or equivalent) may be required for any human subjects research. If you obtained IRB approval, you should clearly state this in the paper. 
        \item We recognize that the procedures for this may vary significantly between institutions and locations, and we expect authors to adhere to the NeurIPS Code of Ethics and the guidelines for their institution. 
        \item For initial submissions, do not include any information that would break anonymity (if applicable), such as the institution conducting the review.
    \end{itemize}

\item {\bf Declaration of LLM usage}
    \item[] Question: Does the paper describe the usage of LLMs if it is an important, original, or non-standard component of the core methods in this research? Note that if the LLM is used only for writing, editing, or formatting purposes and does not impact the core methodology, scientific rigorousness, or originality of the research, declaration is not required.
    %this research? 
    \item[] Answer: \answerNA{} % Replace by \answerYes{}, \answerNo{}, or \answerNA{}.
    \item[] Justification: Our work focuses on visual navigation, which does not involve language models.
    \item[] Guidelines:
    \begin{itemize}
        \item The answer NA means that the core method development in this research does not involve LLMs as any important, original, or non-standard components.
        \item Please refer to our LLM policy (\url{https://neurips.cc/Conferences/2025/LLM}) for what should or should not be described.
    \end{itemize}

\end{enumerate}

%% file: sec/appendix.tex
\section*{\centering{Supplementary Material for \\ \emph{DynaNav: Dynamic Feature and Layer Selection for Efficient Visual Navigation}}}

\section{Overview}
Section~\ref{sec:app_implement} illustrates the hyperparameters and details for training and inference. Section~\ref{sec:app_exp} shows more experimental results. Section~\ref{sec:app_visual} illustrates more visualized results of the saliency heatmaps on observations and goals.

\vspace{-3mm}
\section{Implementation Details}\label{sec:app_implement}
\vspace{-2mm}
For pre-training, we adopt the same parameter settings as ViNT~\cite{shah2023vint} to ensure a fair comparison. A notable difference, however, is that we directly use the encoded features from EfficientNet-B0~\cite{tan2019efficientnet} as tokens to the transformer layer, bypassing the MLP projection used by ViNT to reduce the dimensionality from 1280 to 512. This modification helps save computational resources and time. The input RGB images are resized to a resolution of 85$\times$64, with a batch size of 256. Both ViNT and our models are trained for 100 epochs under these conditions. For fine-tuning, we set the learning rate to 1e-4 and train for 80 epochs, deactivating the warm-up stage during this process.
\input{tables/hyperparam}
\subsection{Hyper-parameter Setting}\label{sec:app_hyperparam}
The detailed hyper-parameter settings for our training and fine-tuning are in Table~\ref{tab:hyper}.

\section{More Experiment Result}\label{sec:app_exp}
\subsection{Results with Mamba Decoder}
We also test the performance of the Mamba~\cite{gu2023mamba} block. Table~\ref{tb:app_mamba} illustrates the results of substituting mamba. As the resolution of our feature maps is not large, the advantage of Mamba~\cite{gu2023mamba} can not be fully explored. On the other hand, the Mamba's~\cite{gu2023mamba} core computing structure - the state space model (SSM) and its high-order recursive calculations- causes its calculation volume to increase rapidly under high-dimensional features. Comparing the $ \text{Sim}(\mathbf{w}_t, \mathbf{w}_t^{\text{gt}})$, the performance using Mamba~\cite{gu2023mamba} blocks is lower than ours. This may result from the fact that Mamba~\cite{gu2023mamba} has advantages on long sequences but may not be able to fully utilize its recursive modeling capabilities on short sequences.
Transformer~\cite{attention} is more suitable for capturing global dependencies. Even if the sequence is short, it can still use the self-attention mechanism to efficiently model the relationship between features.

During the CARLA~\cite{dosovitskiy2017carla} simulation, the model’s predictions are normalized to compute the necessary waypoint offsets. These offsets, combined with the vehicle’s current location, determine the target waypoint. A PID controller is employed to generate control signals based on the target waypoint. To ensure smooth trajectory generation within the CARLA environment, we use an image captured six timestamps ahead of the current observation as the objective, carefully tracking the model’s progress over each run. As ViNT~\cite{shah2023vint} processes the waypoints in relative coordinates, represented as follows:
\begingroup
\begin{equation}
        \mathbf{w}_t = (\mathbf{P}_{t+h} - \mathbf{P}_{t}) \otimes \mathbf{R}(\theta_t),
\end{equation}
\endgroup
where $\mathbf{P}_{t+h}$ and $\mathbf{P}_{t}$ denote the position vectors of the goal and current points in world coordinates, respectively, and $\otimes$ indicates matrix multiplication. $\theta_t$ represents the vehicle's yaw, and $\mathbf{R}$ is the rotation matrix. Thus, the final target point is calculated as: $\mathbf{P}_{t+h} = \mathbf{P}_{t} + \hat{\mathbf{w}}_t \otimes \mathbf{R}(\theta_t)^{\top},$
where $\hat{\mathbf{w}}_t$ is the predicted waypoint offset.

\input{tables/app_mamba}

\subsection{Goal Image Viewpoint Investigation}
In real-world scenarios, goal images often come from diverse sources—such as human-captured photos—and may not exactly align with the agent’s ego-centric view. Understanding how such domain and viewpoint differences impact performance is critical.

To investigate this, we conducted additional experiments in CARLA under three challenging conditions:

\begin{itemize}
\item \textbf{Same location, different angle:} Goal image is taken from the same waypoint but with a camera orientation offset (within $\pm15^\circ$).
\item \textbf{Nearby location, same angle:} Image is captured from a nearby position (within 5 meters), keeping the same orientation.
\item \textbf{Nearby location, different angle:}  Goal is from a nearby waypoint (within 5 meters) and a different orientation.
\end{itemize}

Table~\ref{tb:cam_ablation} illustrates the quantitative results. Our model exhibits graceful degradation as the domain gap increases—i.e., greater viewpoint or positional differences. However, it consistently outperforms the ViNT baseline across all settings, highlighting the robustness and generalization of our approach to goal images with moderate domain shifts.

%To further enhance performance under more significant viewpoint discrepancies, we believe incorporating 3D spatial reasoning—such as depth-aware features or learned spatial priors—would be highly beneficial. We consider this a promising direction for future work and plan to explore it in subsequent research.

\begin{table}[htbp]
\centering
\caption{Comparison of our model and ViNT under varying goal image settings in CARLA Scene A}
\resizebox{0.75\textwidth}{!}{
\begin{tabular}{l l c c}
    \toprule
    \textbf{Setting} & \textbf{Model} & \textbf{Success Rate} & \textbf{FLOPs ($\times 10^9$)} \\
    \midrule
    \multirow{2}{*}{Same Position, Same Angle} 
    & ViNT  & \underline{0.724} & 4.37 \\
    & Ours  & \textbf{0.727}     & \textbf{1.58} \\
    \midrule

    \multirow{2}{*}{Nerby Position, Same Angle} 
    & ViNT  & \underline{0.723} & 4.37 \\
    & Ours  & \textbf{0.725}     & \textbf{1.58} \\
    \midrule

    \multirow{2}{*}{Same Position, Different Angle} 
    & ViNT  & \underline{0.694} & 4.37 \\
    & Ours  & \textbf{0.708}     & \textbf{1.74} \\
    \midrule

    \multirow{2}{*}{Nearby Position, Different Angle} 
    & ViNT  & \underline{0.688} & 4.37 \\
    & Ours  & \textbf{0.691}     & \textbf{1.79} \\
    \bottomrule
\end{tabular}}
\label{tb:cam_ablation}
\end{table}

\subsection{Timestep-wise Consistency}
To investigate the potential timestep-wise inconsistency, we conducted an in-depth analysis on a 700-frame trajectory. We segmented the trajectory into 100-frame intervals and computed the average FLOPs and inference time for each segment, comparing our model against the baseline ViNT. As shown in Table~\ref{tb:app_abltimestep}, our model consistently reduces both computational cost and inference time across all intervals, while maintaining or improving action similarity $\text{Sim}(\mathbf{a}_t,\mathbf{a}_t^{gt})$. In over 96\% of the evaluated trajectories, our approach is more efficient than ViNT without any degradation in navigation accuracy.

These results demonstrate that, despite the dynamic nature of early exiting, our model exhibits stable, consistent, and efficient performance over time in practice.

\begin{table}[htbp]
    \centering
    \caption{Timestep-wise results compared with ViNT}
    \begin{tabular}{l c c c c c c c c}
    \toprule
     & No. of Frame  & 100& 200 & 300 & 400& 500& 600 & 700 \\
    \midrule
     \multirow{3}{*}{ViNT}&FLOPs ($10^9$) & 4.37 & 4.37  & 4.37& 4.37&4.37 &4.37 &4.37\\
    &Avg Time (s) & 0.218 & 0.218 & 0.218 &0.218 & 0.218&0.218 &0.218\\
    &$\text{Sim}(\mathbf{a}_t,\mathbf{a}_t^{gt})$ &94.41 & 94.46 & 94.49 & 94.48&94.52 & 94.51 & 94.49\\
    \midrule
     \multirow{3}{*}{Ours}&FLOPs ($10^9$) & 2.02 & 1.95 & 1.92 & 1.96 &1.85 & 1.93& 1.93\\
    &Avg Time (s) & 0.194 & 0.190 &0.189 &0.191 &0.185 & 0.190 &0.190 \\
    &$\text{Sim}(\mathbf{a}_t,\mathbf{a}_t^{gt})$ &94.76 &  94.88 & 94.92 &94.90 & 94.92 & 94.92 &94.92\\
    \bottomrule
    \end{tabular}
    \label{tb:app_abltimestep}
\end{table}

\subsection{Additional Ablation Study of Constraints}
Our adaptive threshold optimization incorporates three constraints designed to jointly enhance model efficiency across FLOPs, time, and memory usage. To evaluate their individual contributions, we conducted an ablation study by removing each constraint separately.

As shown in Table~\ref{tb:app_reconabl2}, enforcing the FLOPs constraint encourages more frequent layer skipping, effectively reducing inference time and memory consumption. However, removing either the time or memory constraint results in noticeable degradation across all efficiency metrics. This confirms that jointly optimizing all three constraints achieves the best overall performance and balanced resource utilization.

\begin{table}[]
    \centering
    \caption{Ablation Study On RECON Dataset}
    \begin{tabular}{l c c c c}
    \toprule
       Setting  &  FLOPs ($10^9$) & Time (s/traj) & Memory (GB)\\
       \midrule
        Ours  & 1.93 & 0.228 & 13.35 \\
        w/o FLOPs constrain  & 2.84 & 0.291 & 15.62 \\
        w/o Time constrain  & 2.55 & 0.273 & 15.09 \\
        w/o Memory constrain  & 2.16 & 0.255 & 14.75 \\
    \bottomrule
    \end{tabular}
    \label{tb:app_reconabl2}
\end{table}

\subsection{Study on Robustness}
We conducted each navigation trajectory in CARLA 10 times to evaluate the robustness of our method. Table~\ref{tb:app_rand} reports the FLOPs, average execution time, and average successful rate for selected trajectories across these runs. As shown, the results exhibit minimal variance, indicating strong consistency and low randomness. This stability is attributed to the synergy between our feature selector and Bayesian optimization, which together enable adaptive yet reliable behavior across diverse scenarios.

\begin{table}[htbp]
    \centering
    \caption{Results of different separation simulations.}
    \resizebox{\linewidth}{!}{\begin{tabular}{c c c c c c c c c c c c}
    \toprule
     No. of Trajectory  & 1 & 2 & 3 & 4& 5& 6 & 7 & 8& 9 & 10\\
    \midrule
     FLOPs ($10^9$) & 1.91 & 1.93 & 1.92 & 1.91 &1.90 & 1.92& 1.94 &1.93 & 1.93& 1.91\\
    Avg Time (s) & 0.258 & 0.260 &0.258 &0.257 &0.260 & 0.257 &0.262 &0.260 & 0.257 &0.258 \\
    Success Rate & 0.725 & 0.727 &0.727 &0.726 &0.726 & 0.726 &0.728 &0.727 & 0.726 &0.727 \\
    \bottomrule
    \end{tabular}}
    \label{tb:app_rand}
\end{table}

\section{More Visualizations}\label{sec:app_visual}
In this section, we added more visualizations of saliency maps. Such a saliency map helps to identify the interest area after being processed by our proposed feature selector. From Figure~\ref{fig:appviz1} to Figure~\ref{fig:appviz4}, we can tell that the region of interest is not always located in the biggest common object between observation and goal images. The model ``considers'' more spatial information, which results in higher ``attention'' along the target direction. These findings support our claims in Section~\ref{sec:intro} that there is redundant information in the observation and goal. In other words, it proves the rationality of using the proposed feature selector to filter features.
\input{figs/app_viz_1}
\input{figs/app_viz_2}
\input{figs/app_viz_3}
\input{figs/app_viz_4}

%% file: tables/hyperparam.tex
\begin{table}[!h]
    \centering
    \caption{Hyperparameter Settings.}
    \begin{tabular}{l c}
    \toprule
        Hyperparameter & Value  \\
    \midrule
    \textbf{General}&\\  
    \quad   Train Epochs & 100\\
    \quad   Fine-tuning Epochs & 80\\
    \quad   Input Resolution & 85 $\times$ 64\\
    \quad   Training LR &0.0005\\
    \quad   Fine-tuning LR &0.0001\\
    \quad   Warmup Epochs & 3\\
    \quad   Optimizer & AdamW\\
    \quad   LR Scheduler &Cosine Annealing\\
    \quad     Batch Size & 256\\
    \quad $\lambda$ in loss &0.5\\
    
    \textbf{Backbone}&\\  
    \quad     Type & EfficientNet-b0\\
    \quad     Hidden Dim & 1280\\

    \textbf{Data}&\\  
    \quad     Length of past frames & 5\\
    \quad     Length of predicted waypoints & 5\\
    \quad     Max obs-goal distance(meter) & 20\\
    \quad     Min obs-goal distance(meter) & 0\\

    \textbf{Transformer Decoder}&\\  
    \quad     Number of layers    & 4\\
    \quad     Attention Heads & 4\\

    \textbf{Bayesian Optimization}&\\  
    \quad     $\text{Sim}(\mathbf{a}_t, \mathbf{a}_t^{\text{gt}})$ Constraint & 0.950\\
    \quad     $\text{Sim}(\mathbf{w}_t, \mathbf{w}_t^{\text{gt}})$ Constraint &0.960\\
    \quad     FLOPs Constraint ($10^9$) & 2.0\\
    \quad     Time  Constraint (sec) &0.3\\
    \quad     Memory Constraint (GB) & 14 \\
    \quad     Optimization Epochs & 20\\
    \quad       Constraint of Masked Pixels (obs) &2770\\
    \quad       Constraint of Masked Pixels (goal) &3400\\
    \quad       $\xi$ &[0.8,0.5,1.0]\\
      \textbf{CARLA Realted}&\\  
    \quad     Max Speed & 20km/h\\
    \quad     Max Distance & 900m\\
    \quad     Capture Frequency & 4Hz\\
    \bottomrule
    \end{tabular}
    \label{tab:hyper}
\end{table}

%% file: tables/app_mamba.tex
\begin{table}[ht]
    \centering
    \caption{Quantitative Comparison on Benchmarks of ours and Mamba blocks.}
     \resizebox{0.8\linewidth}{!}{\begin{tabular}{c c c c c c c c c}
    \toprule
   Dataset & Method &$ \text{Sim}(\mathbf{w}_t, \mathbf{w}_t^{\text{gt}})$ & FLOPs($10^9$)\\
   
    \midrule
    \multirow{2}{*}{RECON~\cite{shah2023vint}}&Mamba~\cite{gu2023mamba} & 95.09 &4.41\\
    &Ours &96.53  &1.93 \\
    
    \midrule
    \multirow{2}{*}{Go-Stanford~\cite{shah2023gnm}}&Mamba~\cite{gu2023mamba} & 93.34 &4.41\\
    &Ours &93.66  &1.68\\
    
    \midrule
    \multirow{2}{*}{SacSoN~\cite{hirose2023sacson}}&Mamba~\cite{gu2023mamba} & 92.92 &4.41 \\
    &Ours &93.72 &1.68 \\
    
    \midrule
    \multirow{2}{*}{SCAND~\cite{scand}}&Mamba~\cite{gu2023mamba} & 97.28 &4.41\\
    &Ours &97.43 &1.93\\
    \bottomrule
    \end{tabular}}
    \label{tb:app_mamba}
\end{table}

%% file: figs/app_viz_1.tex
\begin{figure*}
    \centering
    \includegraphics[width=1\linewidth]{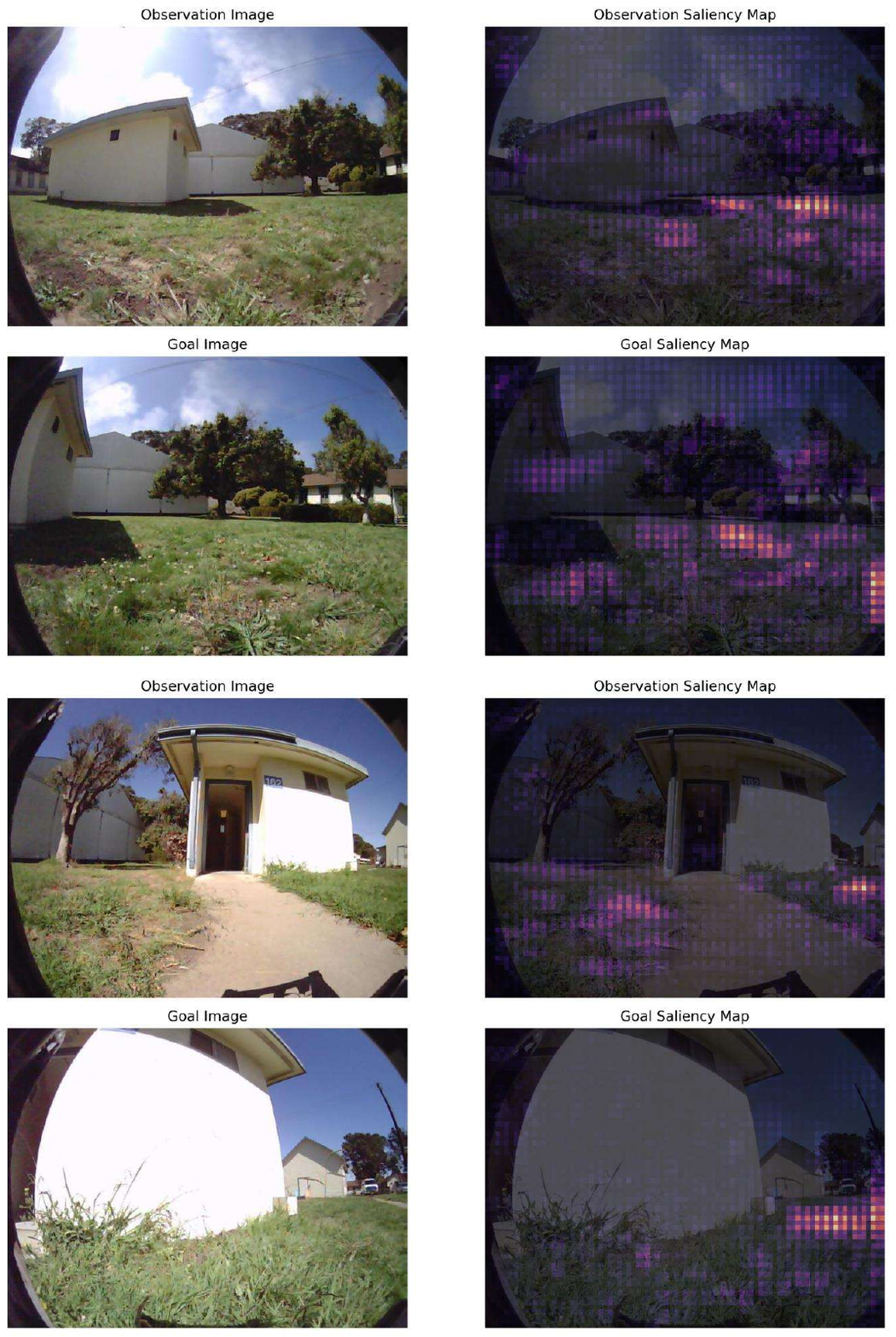}
    \caption{Salieny map of observation and goal images.}
    \label{fig:appviz1}
\end{figure*}

%% file: figs/app_viz_2.tex
\begin{figure*}
    \centering
    \includegraphics[width=0.92\linewidth]{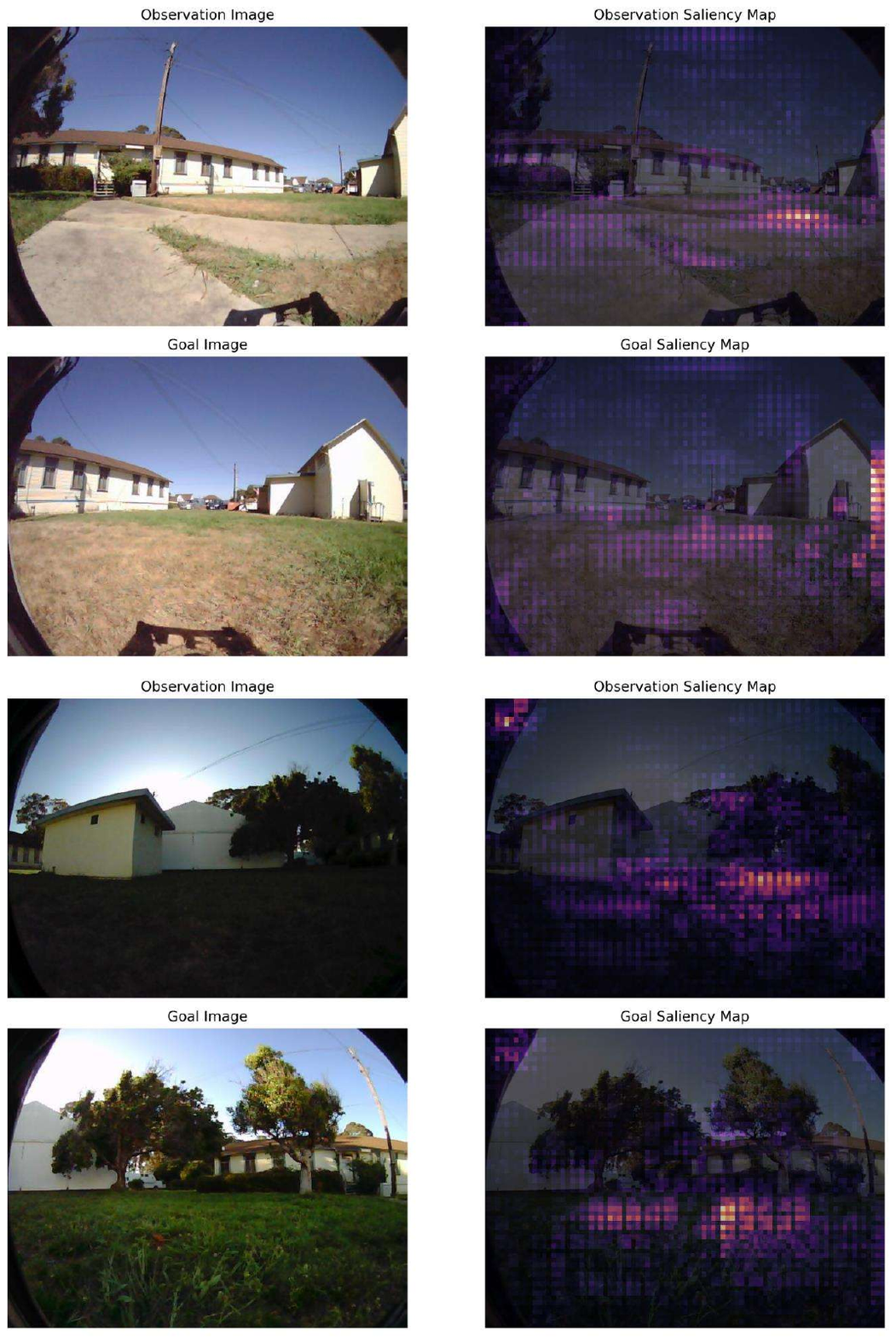}
    \caption{Salieny map of observation and goal images.}
    \label{fig:appviz2}
\end{figure*}

%% file: figs/app_viz_3.tex
\begin{figure*}
    \centering
    \includegraphics[width=0.92\linewidth]{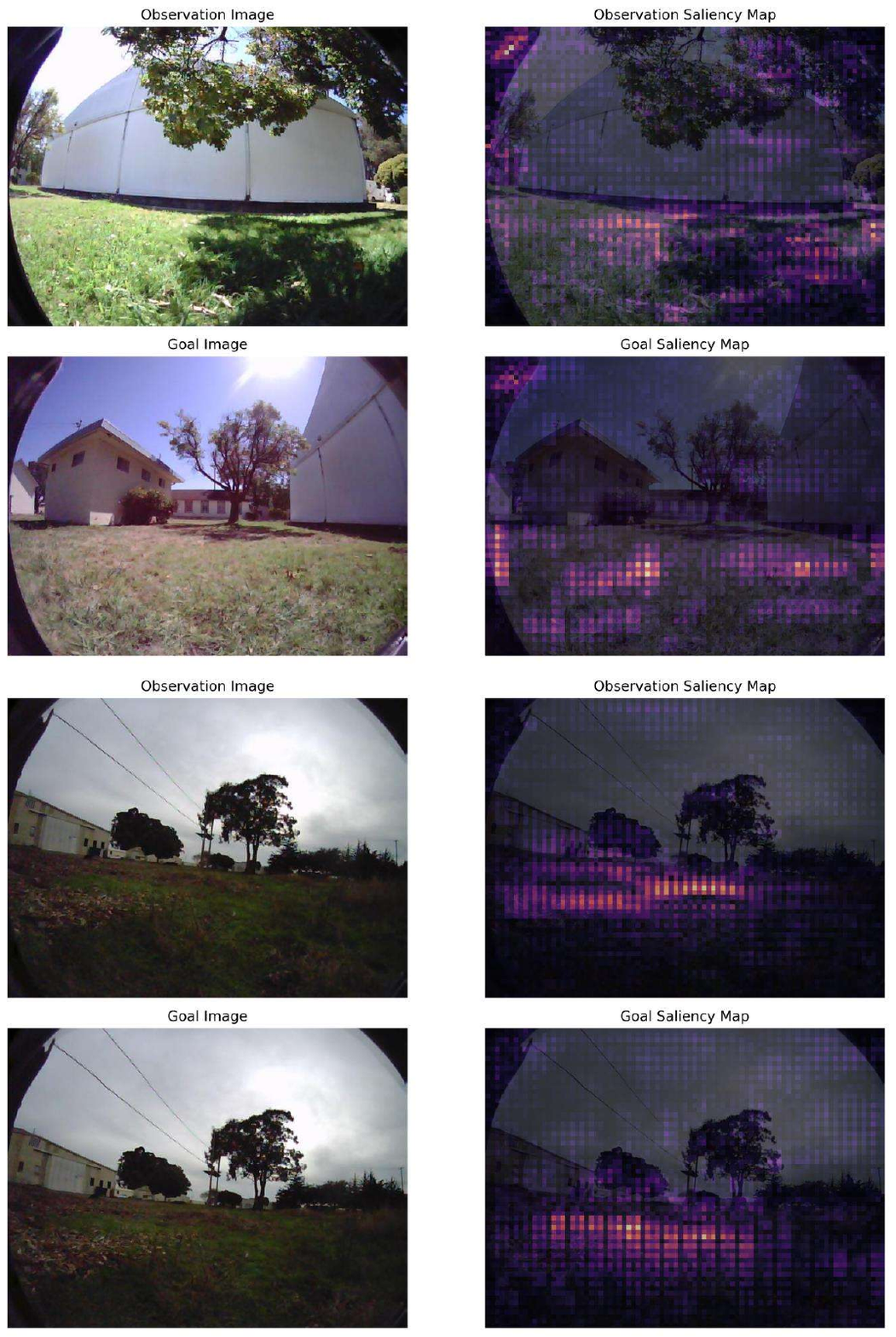}
    \caption{Salieny map of observation and goal images.}
    \label{fig:appviz3}
\end{figure*}

%% file: figs/app_viz_4.tex
\begin{figure*}
    \centering
    \includegraphics[width=0.92\linewidth]{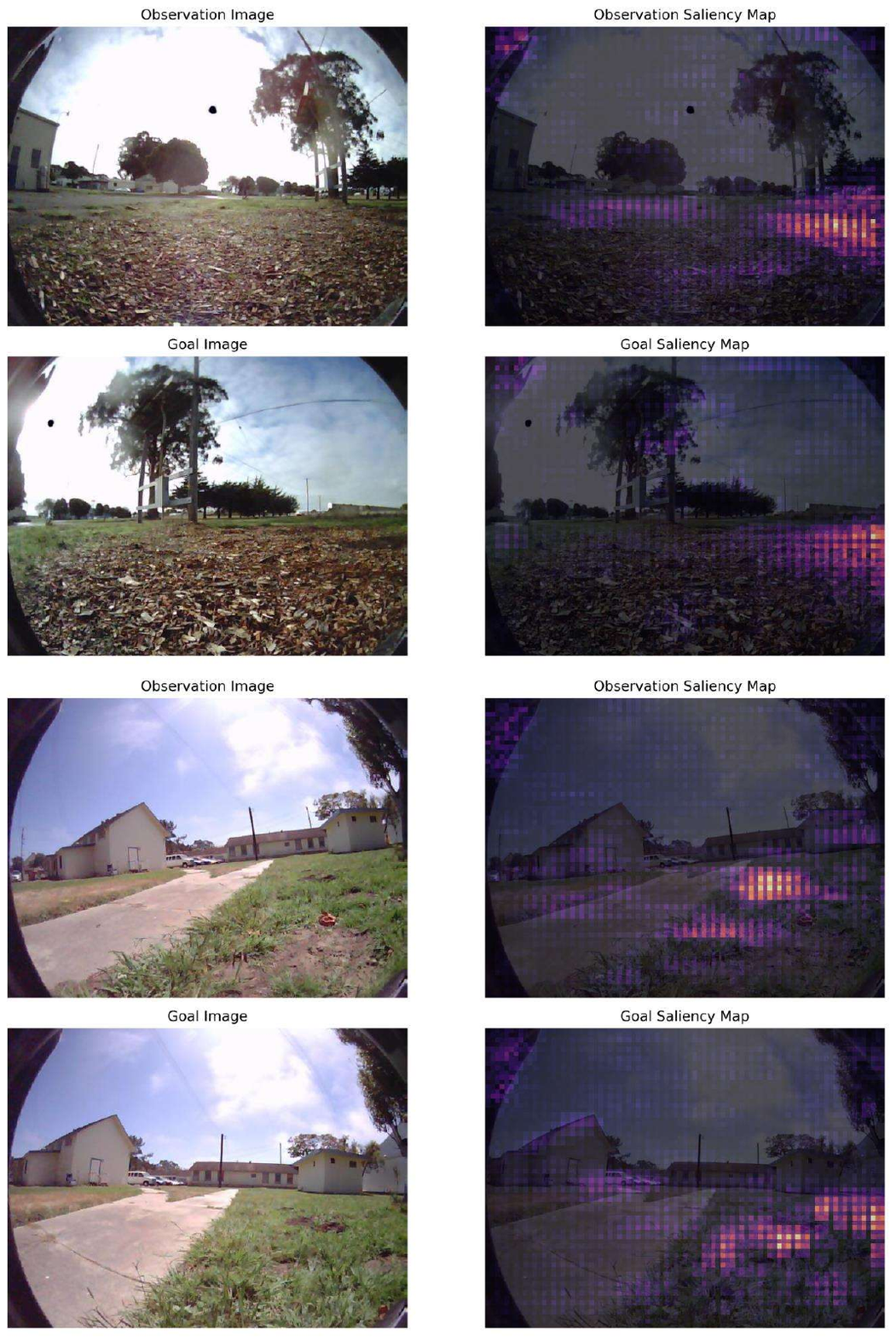}
    \caption{Salieny map of observation and goal images.}
    \label{fig:appviz4}
\end{figure*}